\documentclass[10pt,twocolumn,letterpaper]{article}

\usepackage[pagenumbers]{cvpr}%

\definecolor{oursbg}{HTML}{EAF2F8}%
\definecolor{ourstext}{HTML}{2F5D7E}%

\newcommand{\oursrow}{\rowcolor{oursbg}}
\newcommand{\oursname}[1]{\textcolor{ourstext}{\bfseries #1}}

\usepackage[T1]{fontenc}
\usepackage{amsmath,amssymb}
\usepackage{booktabs}
\usepackage{multirow}
\usepackage{graphicx}
\usepackage{tikz}
\usepackage{xcolor}
\usepackage{adjustbox}
\usepackage{array}

\newcommand{\dreamtourcameraicon}{%
  \begin{tikzpicture}[baseline=-0.4pt,x=1pt,y=1pt]
    \fill[ourstext,rounded corners=0.5pt] (0,0) rectangle (7,4.2);
    \fill[ourstext] (1.5,4.2) -- (2.1,5.2) -- (4.5,5.2)
      -- (5.1,4.2) -- cycle;
    \draw[white,line width=0.5pt] (3.8,2.1) circle (1.2pt);
  \end{tikzpicture}%
}

\providecommand{\cmark}{\textcolor{red!75!black}{\ding{51}}}
\providecommand{\xmark}{\textcolor{green!60!black}{\ding{55}}}

\usepackage{capt-of}

\definecolor{cvprblue}{rgb}{0.21,0.49,0.74}
\usepackage[breaklinks,colorlinks,allcolors=cvprblue]{hyperref}
\usepackage{colortbl}
\usepackage{pifont}

\title{DreamTour: Layout Steering for Training-Free World Exploration}

\author{
Srikumar Sastry$^{1,2}$ \enspace
Meher Gitika Karumuri$^{1}$ \enspace
Chenqi Li$^{1,3}$ \enspace
Sarosij Bose$^{1,4}$ \enspace
Mellon M. Zhang$^{1,5}$ \\
Allen Tu$^{1,6}$ \enspace
Ziyang Long$^{1,8}$ \enspace
Yen-Liang Lin$^{1}$ \enspace
Alexander Schwing$^{1,7}$ \enspace
Nathan Jacobs$^{2}$ \\[5pt]
$^{1}$Amazon \quad
$^{2}$Washington University in St. Louis \quad
$^{3}$University of Oxford \\
$^{4}$University of California, Riverside \quad
$^{5}$Georgia Institute of Technology \\
$^{6}$University of Maryland, College Park \quad
$^{7}$University of Illinois Urbana-Champaign \\
$^{8}$University of California, Los Angeles \\
{\small\url{https://dreamtourls.github.io/}}
}

\hypersetup{
  pdftitle={DreamTour: Layout Steering for Training-Free World Exploration},
  pdfauthor={Srikumar Sastry, Meher Gitika Karumuri, Chenqi Li, Sarosij Bose, Mellon M. Zhang, Allen Tu, Ziyang Long, Yen-Liang Lin, Alexander Schwing, Nathan Jacobs},
  pdfkeywords={World models, video diffusion, camera control, layout steering}
}

\begin{document}
\twocolumn[{
\renewcommand\twocolumn[1][]{#1}
\maketitle
\vspace{-12pt}
\begin{center}
\centering
\includegraphics[width=0.95\textwidth]{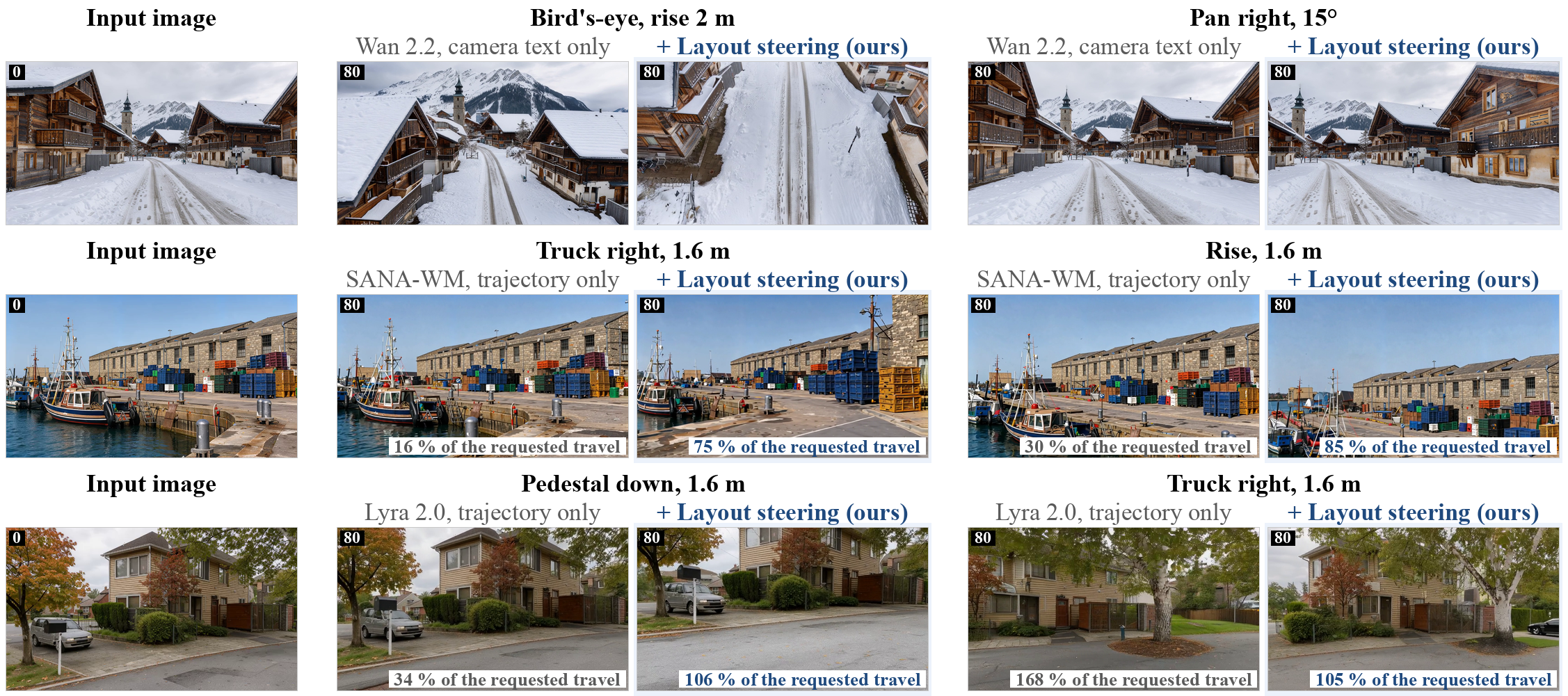}
\vspace{-4pt}
\captionof{figure}{\textbf{DreamTour introduces Layout Steering} for
training-free camera control in video diffusion and world
models~\cite{wan,sanawm,lyra2026}.
Guiding coarse layout during early denoising yields more faithful camera
motion than text or trajectory conditioning alone.}
\label{fig:teaser}
\end{center}
\vspace{5pt}
}]
\begin{abstract}
Video diffusion and world models generate plausible videos
but do not reliably match the requested camera direction and travel
distance. We present DreamTour, a training-free framework that
generates camera tours from a single image with more accurate
control over camera motion. DreamTour first uses a frozen image
inpainting prior to extend the photograph beyond its original field
of view while preserving the input pixels. Depth-based rendering
of this extended scene along the requested camera path produces
approximate target views that serve as spatial guidance.
We introduce Layout Steering, which aligns early video latents
with these views through decaying low-frequency corrections,
leaving later denoising to refine appearance and dynamics.
For longer tours, we autoregressively generate the video segments,
repeating guidance construction from
each segment's final frame to maintain camera control across
the full trajectory. DreamTour improves directional accuracy
and travel calibration on VBench while maintaining competitive
video quality. On WorldScore, it strengthens camera control and
scene consistency across chained movements. Together, these
results demonstrate that training-free spatial guidance enables
more accurate and consistent camera control in pretrained video
diffusion and world models.
\end{abstract}

\section{Introduction}
\label{sec:intro}

Exploring a scene from a single photograph requires controlling where
the camera moves and how far it travels while generating plausible,
spatially consistent unseen content. Existing methods use text, learned
trajectory
conditioning~\cite{motionctrl,cameractrl,gen3c,sanawm}, or training-free
geometric guidance~\cite{camtrol,motioncanvas,camprobe}.
Nonetheless, video generators continue to exhibit \emph{directional bias}, where camera motion
follows learned preferences or scene content rather than the request,
and \emph{distance inconsistency}, where identical requests yield different
travel across seeds and scenes. Motion in the correct direction can still
undershoot or overshoot the requested displacement, even with explicit
trajectory conditioning.

We present \textbf{DreamTour}, a training-free framework that improves
camera direction and travel control in pretrained video diffusion
models~\cite{wan,sanavideo,cogvideox} and camera-conditioned world
models~\cite{sanawm,lyra2026}.
It introduces \textbf{Layout Steering} to align video latents with the
spatial layout of views rendered along the requested path. Decaying
corrections target the low-frequency structure that largely determines camera motion
during early denoising~\cite{ac3d,camprobe}, leaving later steps to refine
appearance~\cite{wang2023diffusion}. Figure~\ref{fig:teaser} illustrates
the improvement in camera control over text or trajectory conditioning
alone.

We construct the spatial guidance for Layout Steering by first
extending the input image beyond its original field of view
using a frozen image inpainting model, while preserving the observed
pixels. We estimate depth and render approximate views of the
extended scene along the requested camera trajectory. During
generation, these views guide the early denoising steps of the
frozen video model, which synthesizes realistic appearance and
dynamics. Camera-conditioned world models additionally retain
their original trajectory inputs. For longer trajectories, we
generate consecutive video segments, each starting from the
previous segment's final frame. We repeat the image outpainting
and guide rendering for each segment, providing an updated
spatial reference that maintains camera guidance across the
full trajectory.

Layout Steering improves directional accuracy and travel calibration on
VBench~\cite{huang2024vbench,huang2025vbench++}, reducing mean direction
error on Wan~2.2~\cite{wan} from $93.8^\circ$ to $32.4^\circ$ while maintaining
competitive video quality. It also improves camera control and the
consistency of scene geometry and appearance across chained movements on
WorldScore~\cite{duan2025worldscore}. Together, these results demonstrate
that DreamTour enables pretrained video diffusion and world models to
generate long-horizon camera tours with more accurate control over camera
direction and travel. Our main contributions are:

\begin{itemize}[itemsep=2pt,topsep=3pt,parsep=0pt]
\item \textbf{Analysis of camera control limitations.}
We use controlled image and seed interventions to measure directional
bias and distance inconsistency in three video diffusion priors and two
world models.

\item \textbf{Training-free Layout Steering.}
We align early video latents with the layout of rendered target views
through decaying low-frequency corrections, improving camera direction
and travel control.

\item \textbf{Scene guidance for camera tours.}
We construct an expanded scene canvas to supply unseen content and a
shared spatial reference, and re-anchor each guide to the generated
handoff view for long-horizon camera tours.
\end{itemize}

\section{Related Work}
\label{sec:related}

Video diffusion models like Wan~\cite{wan}, SANA-Video~\cite{sanavideo}, and CogVideoX~\cite{cogvideox} generate realistic videos from text or images but offer limited control over camera motion.
Learning-based methods like TrajectoryCrafter~\cite{trajectorycrafter} and ReCamMaster~\cite{recammaster} add camera-trajectory conditioning during training~\cite{motionctrl,cameractrl}.
Training-free methods steer pretrained video models at inference time by warping scene content or video latents along a target path~\cite{motioncanvas,nvssolver,viewextrapolator,latentreframe}.
CamTrol~\cite{camtrol} renders a point cloud from the requested views, while CamProbe~\cite{camprobe} geometrically resamples video latents.
Section~\ref{sec:bias} extends CamProbe's analysis to spontaneous and requested camera motion.

Video world models extend camera control to interactive generation by responding to user actions or camera trajectories over time~\cite{agarwal2026cosmos,genie3,matrixgame}.
SANA-WM~\cite{sanawm} produces minute-scale videos from camera paths.
GEN3C~\cite{gen3c} uses renderings from a persistent 3D cache, while Voyager~\cite{voyager} jointly predicts RGB and depth to maintain the scene over long trajectories.
Related single-image methods like WonderJourney~\cite{yu2024wonderjourney} and WonderWorld~\cite{wonderworld} expand scenes beyond the input view.
Despite these approaches, video diffusion models and world models still cannot reliably control camera direction and travel distance.

Training-free diffusion guidance can constrain spatial composition and motion in videos.
MultiDiffusion~\cite{multidiffusion} fuses overlapping generations, while other methods coordinate views or preserve observed pixels~\cite{synctweedies,stochsync,repaint}.
Sampling can be guided by matching low-frequency structure, refining the initial noise, or using early teacher predictions~\cite{ilvr,freeinit,videoguide}.
Go-with-the-Flow~\cite{gowiththeflow} warps the initial noise to impose a target motion.
AC3D~\cite{ac3d} shows that camera motion emerges in low-frequency structure early in denoising.
DreamTour exploits this property by aligning early, low-frequency video latents with views rendered from an outpainted canvas along the target path.
Later steps synthesize parallax, disocclusions, and high-frequency detail.

\section{Camera Control Limitations}
\label{sec:bias}

We first analyze whether video models follow requested camera directions and produce consistent travel distances. We evaluate the video diffusion models Wan~2.2~\cite{wan}, SANA-Video~2.0~\cite{sanavideo}, and CogVideoX~1.5~\cite{cogvideox}, and the trajectory-conditioned world models SANA-WM~\cite{sanawm} and Lyra~2.0~\cite{lyra2026} on 24 synthetic frames from FLUX.3~\cite{flux3}. Repeated conditions use up to five seeds (Appendix~\ref{app:bias}). Camera motion is estimated from accumulated median optical flow and flow divergence. Table~\ref{tab:bias} summarizes the results.

\noindent \textbf{Directional bias.}
We evaluate five pretrained models under six camera conditions:
no instruction, stationary, left, right, forward, and rising.
We also mirror the input images to distinguish model-specific
travel preferences from scene-dependent behavior. Without camera
instructions, SANA-Video exhibits substantial drift (30.5 px)
that rarely reverses when the input is mirrored (4.2\%),
suggesting a learned directional preference. CogVideoX also
drifts (12.6 px), but reverses its travel direction in 45.8\%
of mirrored cases, indicating greater sensitivity to scene
content. Under opposing left/right requests, Wan~2.2 follows
both directions in 87.5\% of cases, while SANA-WM and Lyra~2.0
succeed in all cases. In contrast, neither SANA-Video nor
CogVideoX reliably follows both requests.

\noindent \textbf{Scene dependence.}
Camera travel also depends on the starting image's composition.
Wan~2.2, CogVideoX, and Lyra~2.0 travel away from the nearer
side of the scene in 70.8\%, 66.7\%, and 66.7\% of cases,
respectively. CogVideoX shows the strongest correlation
between travel direction and the position of nearby scene
content ($\rho=-0.48$). Both CogVideoX and Lyra~2.0 also
exhibit correlations between zoom and openness near the
image centre ($\rho=0.50$). Across random seeds, the
direction of travel remains relatively stable
(73.3--98.6\% agreement), suggesting that scene composition
and learned travel preferences systematically influence
camera movement.

\noindent \textbf{Distance inconsistency.}
Following the correct direction does not guarantee consistent
travel distance. We measure signed truck displacement along the
requested direction, where negative values indicate opposite
travel. Wan~2.2 and SANA-WM exhibit mean displacements of
60.5 and 46.8 pixels, respectively. In contrast, SANA-Video
and CogVideoX show near-zero mean displacements ($-0.6$ and
0.5 px), with opposite average travel directions and substantial
variation across clips. Changing only the random seed produces
26.6\% and 28.7\% travel variation for Wan~2.2 and SANA-WM,
respectively. Only 21.2\% and 24.6\% of their clips fall
within $\pm25\%$ of median travel, compared with 0.0\%
and 1.1\% for SANA-Video and CogVideoX. Lyra~2.0 is more
consistent across seeds (3.8\%) but exhibits the largest mean
displacement (143.7 px) and substantial across-clip variation
(57.5 px). Despite trajectory conditioning, SANA-WM travels
$2.2\times$ farther along a trajectory than its mirrored
counterpart, whereas Lyra~2.0 maintains a ratio of 1.0.
These findings motivate spatial guidance that constrains
both camera direction and travel magnitude.

\begin{figure*}[t]
  \centering
  \includegraphics[width=\textwidth]{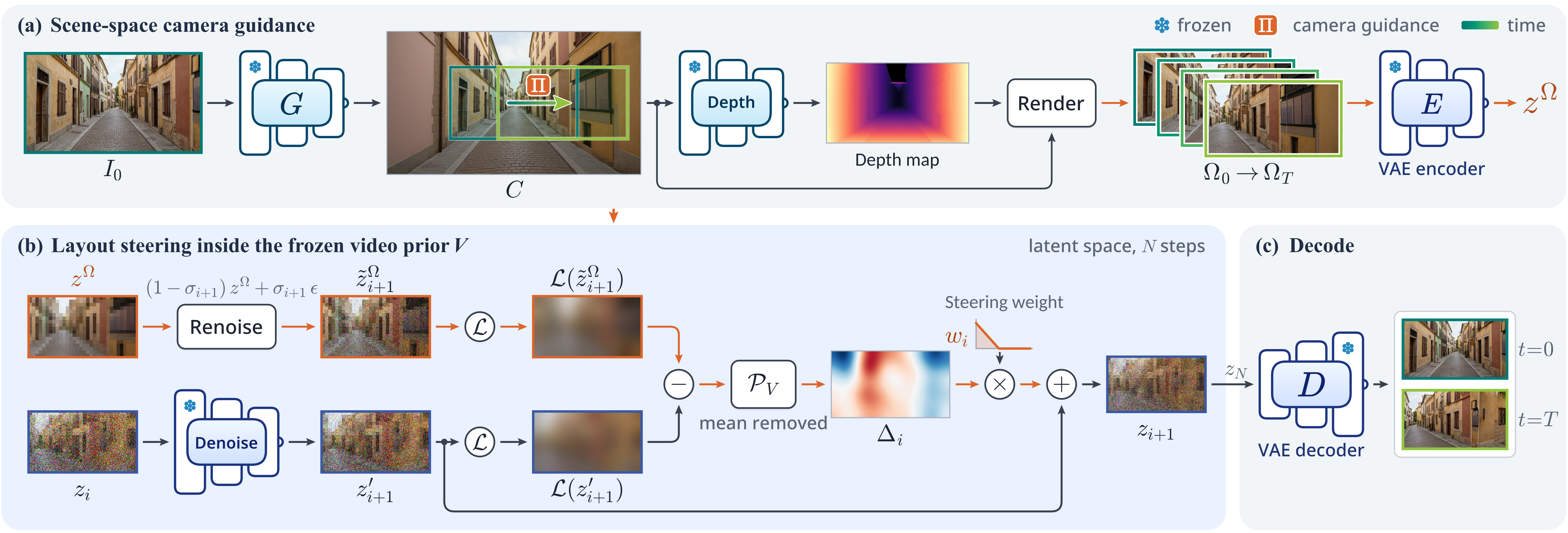}
  \caption{\textbf{Overview of DreamTour.}
  (a) A frozen inpainting model expands the input image into a canvas.
  We render views along $\Pi$ using depth from $\pi^3$X~\cite{pi3} and
  encode them as the guide latent.
  (b) Decaying low-frequency corrections steer the frozen video prior's
  latent for the first $K$ of $N$ denoising steps, followed by $N-K$
  steps without steering.
  (c) The final latent is decoded into a coherent video.}
  \label{fig:method}
\end{figure*}

\section{Method}

\label{sec:method}

\textbf{DreamTour} generates scene-consistent video by making the
requested camera direction and travel explicit in spatial layouts.
It takes an input image $I_0$, a caption $c$, and a camera trajectory
$\Pi$ specifying translations in metres and rotations in degrees
relative to the input camera. As Figure~\ref{fig:method} shows, a
frozen image inpainting model $G$ expands the scene beyond $I_0$
into a shared canvas. Rendering this canvas along $\Pi$ produces a
guide that steers the low-frequency structure of a frozen video
generator $V$ during early denoising. The guide constrains camera
motion and coarse structure, while $V$ supplies realistic dynamics
and fine detail. This training-free design applies to both
image-to-video diffusion models and camera-conditioned world models.

\subsection{Scene-Space Camera Guidance}
\label{sec:canvas}

Camera motion reveals regions beyond the input view, requiring
plausible and spatially consistent scene completion. We construct
an expanded scene canvas $C$ using FLUX.1 Fill~\cite{fluxfill} as the
frozen inpainting model $G$. Centering the input image $I_0$ on $C$
leaves horizontal and vertical margins for newly revealed content.
A single outpainting pass conditioned on the caption $c$ fills
these margins while preserving the input pixels exactly. This
separates scene expansion from video synthesis and gives successive
views a shared spatial reference.

\begin{table}[t]
  \centering
  \caption{\textbf{Correct direction does not guarantee consistent travel and vice versa.}
  We evaluate directional control, the influence of scene
  composition, and travel consistency across video diffusion
  and world models. Mirroring tests whether flipping the input
  reverses camera travel. Scene correlations ($\rho$) relate
  travel to the starting image's composition. Travel consistency
  is measured relative to each model's median displacement along the requested direction.}
  \label{tab:bias}

  \scriptsize
  \setlength{\tabcolsep}{3.5pt}
  \renewcommand{\arraystretch}{1.20}
  \providecolor{oursbg}{RGB}{237, 240, 245}
  \providecolor{ourstext}{RGB}{31, 73, 125}

  \resizebox{\columnwidth}{!}{%
  \begin{tabular}{@{}lccccc@{}}
    \toprule
    & \multicolumn{3}{c}{\textbf{Video Diffusion Models}}
    & \multicolumn{2}{c}{\textbf{World Models}} \\
    \cmidrule(lr){2-4}
    \cmidrule(l){5-6}

    \textbf{Metric}
    & \textbf{Wan~2.2}
    & \textbf{SANA-V2}
    & \textbf{CogVideoX}
    & \textbf{SANA-WM}
    & \textbf{Lyra~2.0} \\
    \midrule

    \rowcolor{oursbg}
    \multicolumn{6}{@{}l}{
      \textcolor{ourstext}{
        \textbf{Following the requested direction}}} \\

    Travel without instruction
      & 1.7\,px & 30.5\,px & 12.6\,px
      & 1.1\,px & 0.1\,px \\

    Travel reverses on input image flipping
      & 54.2\% & 4.2\% & 45.8\%
      & 8.3\% & 4.2\% \\

    Both left/right requests followed
      & 87.5\% & 0.0\% & 0.0\%
      & 100.0\% & 100.0\% \\

    \addlinespace[2pt]

    \rowcolor{oursbg}
    \multicolumn{6}{@{}l}{
      \textcolor{ourstext}{
        \textbf{Influence of scene composition}}} \\

    Travel away from nearer side
      & 70.8\% & 41.7\% & 66.7\%
      & 45.8\% & 66.7\% \\

    Travel vs. nearer-side position ($\rho$)
      & $-0.22$ & 0.05 & $-0.48$
      & $-0.13$ & $-0.27$ \\

    Zoom vs. centre openness ($\rho$)
      & $-0.17$ & 0.21 & 0.50
      & -- & 0.50 \\

    Same travel direction across seeds
      & 73.3\% & 90.0\% & 75.0\%
      & 94.2\% & 98.6\% \\

    \addlinespace[2pt]

    \rowcolor{oursbg}
    \multicolumn{6}{@{}l}{
      \textcolor{ourstext}{
        \textbf{Consistency of travel distance}}} \\

    Mean truck displacement
      & 60.5\,px & $-0.6$\,px & 0.5\,px
      & 46.8\,px & 143.7\,px \\

    Travel std. dev. across clips
      & 49.1\,px & 49.8\,px & 39.0\,px
      & 37.7\,px & 57.5\,px \\

    Travel variation across seeds
      & 26.6\% & -- & --
      & 28.7\% & 3.8\% \\

    Within $\pm25\%$ of median travel
      & 21.2\% & 0.0\% & 1.1\%
      & 24.6\% & 59.7\% \\

    Travel ratio for mirrored paths
      & -- & -- & --
      & 2.2 & 1.0 \\

    \bottomrule
  \end{tabular}%
  }
\end{table}

To render the guide, we estimate canvas depth with the frozen
monocular depth estimator $\pi^3$X~\cite{pi3}, normalize it by its median,
and treat sky regions as infinitely distant. We then rescale the
normalized depth once using the median metric depth predicted for the
input image by Depth Anything~3~\cite{da3}, so the scene geometry and
commanded translations share a metric scale before video generation.
Unprojecting the canvas $C$ with the calibrated depth and camera
intrinsics yields a textured 3D point cloud.

\begin{figure}[t]
  \centering
  \input{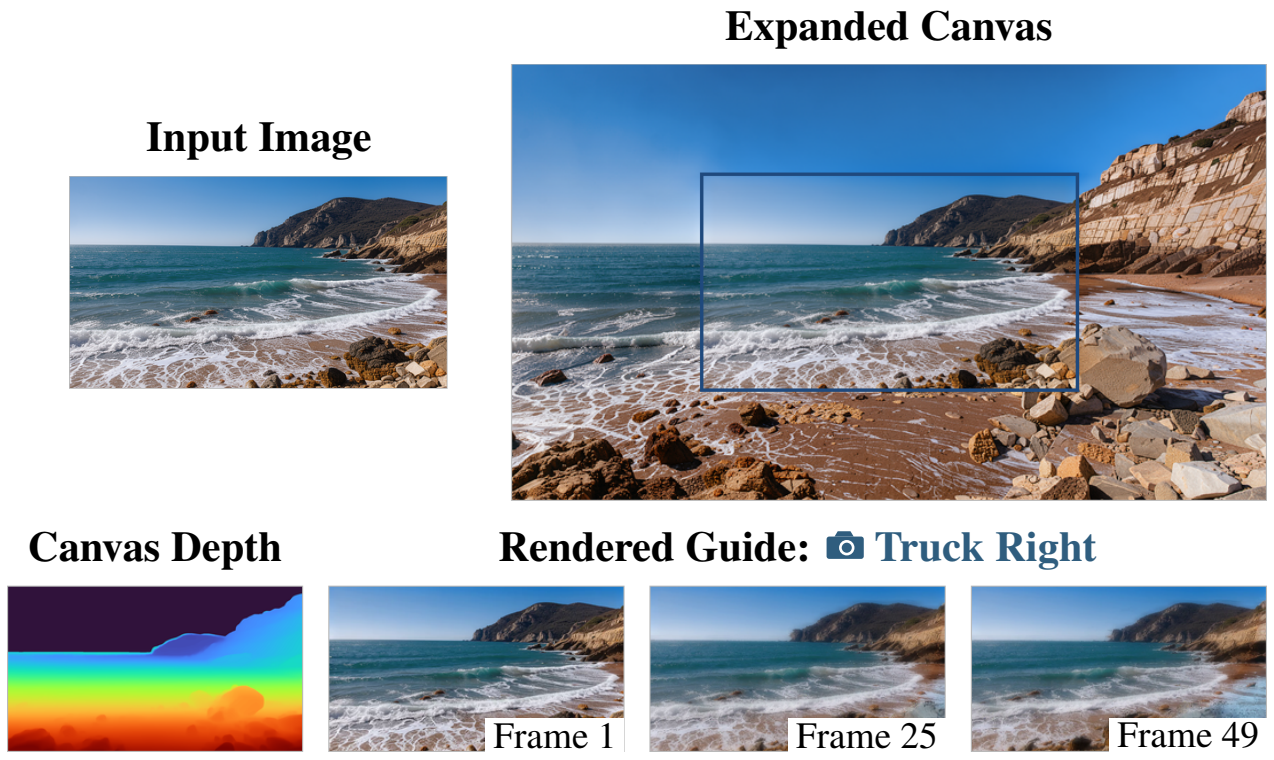}
  \caption{\textbf{Canvas and guide construction.}
  A frozen inpainting prior expands the input image while preserving
  its pixels exactly. Estimated depth lifts the canvas into a textured
  3D point cloud, rendered along the requested camera path to form a
  guide video.}
  \label{fig:canvas_guide}
\end{figure}

For a camera trajectory $\Pi=\{P_t\}_{t=0}^{T-1}$ with $T$ frames,
$P_t$ denotes the camera pose at frame $t$ relative to the input
camera. At each pose, we render the point cloud using z-buffered
splatting followed by coarse hole filling to obtain a guide frame
$\Omega_t$. This rendering supports translations, rotations, orbits,
and bird's-eye paths. Figure~\ref{fig:canvas_guide} shows the resulting
guide video $\Omega=\{\Omega_t\}_{t=0}^{T-1}$, which captures the direction
and magnitude of image-space motion and the content entering each view.
The frozen video prior synthesizes realistic appearance and dynamics.

\subsection{Low-Frequency Layout Steering}
\label{sec:steering}

The guide provides the desired layout, but copying its pixels would
also transfer reprojection artifacts and errors from incomplete
geometry. Early denoising establishes global structure and camera
motion, while later steps refine detail~\cite{ac3d,camprobe}.
We introduce \textbf{Layout Steering} to guide only the low-frequency
structure of early video latents toward
the encoded guide $z^\Omega=E_V(\Omega)$, where $E_V$ is the encoder
of $V$'s variational autoencoder. A spatial
Gaussian low-pass filter $\mathcal{L}$ acts independently on each
latent frame and channel, with a standard deviation of three latent
cells.

At denoising step $i$, $V$ updates the video latent $z_i$ to
$z'_{i+1}$ before steering. We steer the first $K$ of $N$ denoising
steps, using $40\%$ of the schedule by default. For $i=0,\ldots,K-1$, we add noise
to the guide at the current noise level $\sigma_{i+1}\in[0,1]$ using
standard Gaussian noise $\epsilon$ of the same shape as $z^\Omega$.
The steering update uses model-dependent normalization $\mathcal{P}_V$
and weight $w_i$:
\vspace{-4pt}
\begin{equation}
\begin{aligned}
\tilde z_{i+1}^{\Omega}
&=(1-\sigma_{i+1})z^\Omega
+\sigma_{i+1}\epsilon,\\
z_{i+1}&=z'_{i+1}+w_i\,\mathcal{P}_V\!\left[
\mathcal{L}(\tilde z_{i+1}^{\Omega})-\mathcal{L}(z'_{i+1})\right].
\end{aligned}
\label{eq:layout}
\end{equation}
\noindent
Here $\tilde z_{i+1}^{\Omega}$ is the noised guide and $z_{i+1}$ is
the steered video latent. Matching the noise levels makes the guide
and video latent comparable. The weight $w_i=w[1-(i+1)/K]$ decays
linearly to zero, where $w$ controls the guidance strength, and we use the frozen sampler without steering for
the remaining $N-K$ steps. Appendix~\ref{app:guidance} interprets this update as a spatially filtered gradient step.

Low-frequency corrections can also transfer the guide's color and
exposure. For most of the evaluated video priors~\cite{nvidia2025cosmos,
hunyuanvideo15,wan,cogvideox}, we use a mean-free correction:
\vspace{-4pt}
\begin{equation}
\begin{aligned}
\delta_i&=\mathcal{L}(\tilde z_{i+1}^{\Omega})
-\mathcal{L}(z'_{i+1}),\\
\mathcal{P}_V(\delta_i)
&=\delta_i-\mu_{\mathrm{spatial}}(\delta_i).
\end{aligned}
\label{eq:tone}
\end{equation}
where $\delta_i$ is the low-frequency mismatch before normalization
and $\mu_{\mathrm{spatial}}$ averages over height and width for
each latent frame and channel. This preserves the latent's per-frame
channel means while transferring spatial variation. For SANA-Video~\cite{sanavideo}
and SANA-WM~\cite{sanawm}, whose compressed latents also carry layout
in their channel statistics, we retain the full correction by setting
$\mathcal{P}_V(\delta_i)=\delta_i$.

The frozen prior supplies realistic dynamics, parallax, disocclusion,
and fine detail. For camera-conditioned world models, we retain the
original trajectory input and add an image-space reference for the
requested motion.

\subsection{Long-Horizon Camera Tours}
\label{sec:chaining}

To extend camera tours beyond a single generation window, we divide
$\Pi$ into consecutive segments and use each window's final frame as
the input to the next. Reusing the original guide can compound
discrepancies between the prescribed and generated handoff views.
We therefore \emph{re-anchor} the canvas by expanding the handoff frame
with $G$ and rendering a fresh guide for layout steering along the
next segment. Keeping the handoff resolution fixed prevents artificial
zoom from repeated cropping and rescaling.

\begin{table*}[t]
  \centering
  \caption{\textbf{Camera control and video quality under the CamProbe~\cite{camprobe} protocol.}
  We evaluate eight camera motions on VBench~\cite{huang2024vbench}
  using Depth Anything~3~\cite{da3} poses.
  \mbox{RPE-T} and \mbox{RPE-R} measure raw relative translation and rotation
  errors without trajectory alignment or scale fitting. Dir.\ measures mean angular error in net motion.
  Travel and rotation errors report median $|r-1|$,
  where $r$ is realized motion along the requested direction. Quality metrics assess imaging,
  smoothness, and background consistency. The training column indicates whether the method requires camera-control training beyond its base video prior. Layout Steering adds no training, including when applied to already camera-conditioned world models.}
  \label{tab:camprobe}

  \footnotesize
  \setlength{\tabcolsep}{4.5pt}
  \renewcommand{\arraystretch}{1.18}
  \resizebox{\linewidth}{!}{
  \begin{tabular}{@{}llccccccccc@{}}
    \toprule
    & & &
    \multicolumn{2}{c}{\textbf{Camera Control} $\downarrow$}
    & \multicolumn{3}{c}{\textbf{Camera Aim \& Displacement} $\downarrow$}
    & \multicolumn{3}{c}{\textbf{Video Quality} $\uparrow$} \\
    \cmidrule(lr){4-5}
    \cmidrule(lr){6-8}
    \cmidrule(l){9-11}

    \multirow[b]{-2}{*}{\textbf{Base Video Prior}}
    & \multirow[b]{-2}{*}{\textbf{Method}}
    & \multirow[b]{-2}{*}{\shortstack{\textbf{New Control}\\\textbf{Training}}}
    & \textbf{RPE-T (m)}
    & \textbf{RPE-R ($^\circ$)}
    & \textbf{Dir.\ ($^\circ$)}
    & \textbf{Travel Err.}
    & \textbf{Rot. Err.}
    & \textbf{Imaging}
    & \textbf{Smoothness}
    & \textbf{Consistency} \\

    \midrule

    & Gen3C~\cite{gen3c} & \cmark
      & 0.044 & 0.67 & \textbf{5.2} & 0.36 & \textbf{0.05} & 70.84 & 99.21 & \textbf{97.09} \\

    \rowcolor{oursbg}
    \cellcolor{white}\multirow{-2}{*}{Cosmos-Predict1 7B~\cite{nvidia2025cosmos}}
      & \textcolor{ourstext}{\textbf{+ Layout Steering (Ours)}} & \xmark
      & \textbf{0.038}
      & \textbf{0.59} & 44.3 & \textbf{0.32} & 0.44%
      & \textbf{70.97}
      & \textbf{99.24}
      & 96.97 \\

    \midrule

    & TrajectoryCrafter~\cite{trajectorycrafter} & \cmark
      & 0.058 & \textbf{0.84} & 37.6 & 0.65 & 0.51
      & \textbf{70.90} & \textbf{99.20} & 96.14 \\

    \rowcolor{oursbg}
    \cellcolor{white}\multirow{-2}{*}{CogVideoX-Fun 5B~\cite{cogvideox}}
      & \textcolor{ourstext}{\textbf{+ Layout Steering (Ours)}} & \xmark
      & \textbf{0.045}
      & 0.88 & \textbf{32.6} & \textbf{0.59} & \textbf{0.48}%
      & 69.29
      & 98.79
      & \textbf{96.21} \\

    \midrule

    & ReCamMaster~\cite{recammaster} & \cmark
      & 0.034
      & 0.58 & \textbf{26.5} & 0.63 & \textbf{0.30}
      & 70.81
      & \textbf{99.24}
      & 96.03 \\

    \rowcolor{oursbg}
    \cellcolor{white}\multirow{-2}{*}{Wan~2.1 1.3B~\cite{wan}}
      & \textcolor{ourstext}{\textbf{+ Layout Steering (Ours)}} & \xmark
      & \textbf{0.031}
      & \textbf{0.49} & 29.0 & \textbf{0.55} & 0.42%
      & \textbf{71.08}
      & 98.84
      & \textbf{96.42} \\

    \midrule

    & Camera Text Alone & \xmark
      & 0.053 & 1.18 & 93.8 & 1.01 & 1.02
      & \textbf{71.49} & 98.15 & 95.49 \\

    & + CamTrol~\cite{camtrol} & \xmark
      & 0.045 & 0.88 & 78.0 & 1.00 & 0.95
      & 70.93 & 99.12 & 98.00 \\

    & + CamProbe~\cite{camprobe} & \xmark
      & 0.065 & 0.77 & 70.7 & 0.99 & 0.89
      & 68.58 & 99.08 & 96.03 \\

    \rowcolor{oursbg}
    \cellcolor{white}\multirow{-4}{*}{Wan~2.2 5B~\cite{wan}}
      & \textcolor{ourstext}{\textbf{+ Layout Steering (Ours)}} & \xmark
      & \textbf{0.040}
      & \textbf{0.74} & \textbf{32.4} & \textbf{0.68} & \textbf{0.39}
      & 71.33
      & \textbf{99.76}
      & \textbf{98.27} \\

      \midrule

    & SANA-WM~\cite{sanawm} & \cmark
      & 0.039 & 0.60 & 70.0 & 1.00 & 1.00
      & 71.66 & 99.29 & \textbf{97.95} \\

    \rowcolor{oursbg}
    \cellcolor{white}\multirow{-2}{*}{SANA-Video~1.0 2B~\cite{sanavideo}}
      & \textcolor{ourstext}{\textbf{+ Layout Steering (Ours)}} & \xmark
      & \textbf{0.032}
      & \textbf{0.41}
      & \textbf{37.3}
      & \textbf{0.68} & \textbf{0.39}
      & \textbf{71.73}
      & \textbf{99.40}
      & \textbf{97.95} \\

      \midrule

    & Lyra~2.0~\cite{lyra2026} & \cmark
      & 0.067 & 0.70 & 56.0 & 2.12 & 1.06
      & 71.85 & 99.10 & \textbf{95.30} \\

    \rowcolor{oursbg}
    \cellcolor{white}\multirow{-2}{*}{Wan~2.1 14B~\cite{wan}}
      & \textcolor{ourstext}{\textbf{+ Layout Steering (Ours)}} & \xmark
      & \textbf{0.048}
      & \textbf{0.64} & \textbf{55.8} & \textbf{1.64} & \textbf{0.64}
      & \textbf{71.89}
      & \textbf{99.19}
      & 94.98 \\

    \bottomrule
  \end{tabular}}
\end{table*}

\subsection{Measuring Camera Motion}
\label{sec:motion_metrics}

Accurate scene exploration requires reaching the intended viewpoint, not merely matching the shape of a camera trajectory. We therefore complement raw, unaligned relative pose error (RPE)~\cite{camprobe} with metrics that explicitly evaluate net camera motion.

We jointly reconstruct each generated clip and its trajectory-rendered
guide in a shared coordinate system and scale with the metric
checkpoint\footnote{\scriptsize https://huggingface.co/depth-anything/DA3NESTED-GIANT-LARGE-1.1}
of Depth Anything~3~\cite{da3}. The model estimates poses and intrinsics
from image frames alone using OpenCV camera conventions. Commanded
start and target camera centers define the requested displacement
$\mathbf{t}_{\mathrm{req}}=\mathbf{C}_{\mathrm{target}}-\mathbf{C}_{\mathrm{start}}$
in the initial camera frame. We set scale from the input view's
estimated median depth alone, without post-hoc alignment.

Let $\mathbf{t}$ denote the estimated first-to-last camera translation, expressed in the same metric frame as $\mathbf{t}_{\mathrm{req}}$. We define the signed \emph{motion ratio} as $r=(\mathbf{t}\cdot\mathbf{t}_{\mathrm{req}})/\|\mathbf{t}_{\mathrm{req}}\|^2$, measuring realized travel along the requested direction. Here, $r=1$ indicates the requested displacement, $r=0$ indicates no net travel along the requested axis, and $r<0$ indicates reversed motion. For rotations, we analogously project net rotation onto the requested axis. We report \emph{direction error} as the angular deviation from the requested motion, together with \emph{travel and rotation errors} as the median $|r-1|$ across clips. Table~\ref{tab:camprobe} reports these metrics alongside raw RPE and VBench~\cite{huang2024vbench,huang2025vbench++} quality scores.

\section{Experiments}
\label{sec:experiments}

We evaluate DreamTour's camera control, video quality, and scene
consistency within single clips and across chained camera tours.
All experiments use publicly released checkpoints with frozen model
weights. Since video priors differ in latent representations and denoising dynamics, we tune the steering hyperparameters separately for each model using 24 synthetic FLUX.3 scenes. Table~\ref{tab:ablation} presents ablations on Wan~2.2,
and Appendix~\ref{app:ablations} reports additional ablations for the world models.

\begin{figure*}[t]
  \centering
  \includegraphics[width=\textwidth]{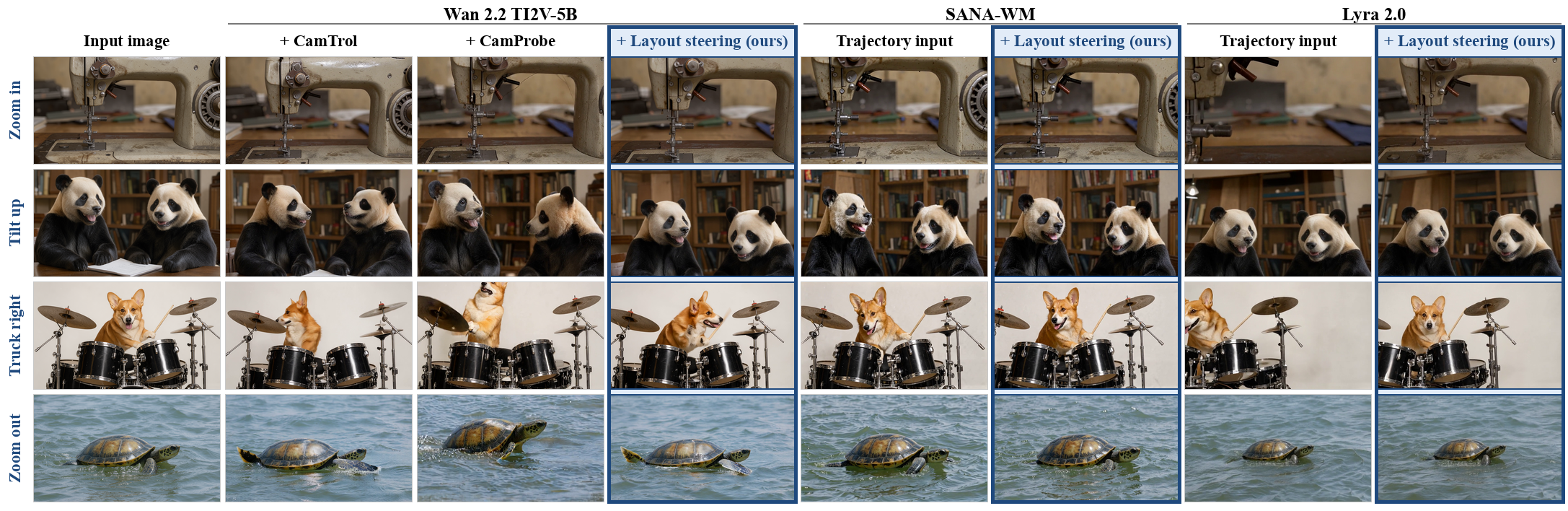}
  \caption{\textbf{Camera control under the CamProbe~\cite{camprobe} protocol.}
  On VBench~\cite{huang2024vbench}, Layout Steering improves camera
  following over CamTrol~\cite{camtrol} and CamProbe~\cite{camprobe} on
  Wan~2.2~\cite{wan} and trajectory conditioning alone~\cite{sanawm,lyra2026}
  while revealing plausible new scene content.}
  \label{fig:lay_results}
\end{figure*}

\subsection{Camera Control Across Generative Models}
\label{sec:exp_wm}

Following the CamProbe~\cite{camprobe} protocol, we evaluate eight camera
motions across 93 VBench~\cite{huang2024vbench,huang2025vbench++} prompts,
yielding up to 744 single-window clips per method. Each clip contains
49 frames at 16\,fps, typically evaluated at $832\times480$ resolution.
Appendix~\ref{app:camprobe} details resolution and completion-count exceptions. The SANA-Video checkpoint listed as SANA-WM's base prior is distinct from the SANA-Video~2.0 5B checkpoint in the camera-bias probe.
Table~\ref{tab:camprobe} compares text-only and learned camera control
alongside training-free guidance from CamTrol~\cite{camtrol} and CamProbe.

On Wan~2.2, Layout Steering reduces mean direction error from
$93.8^\circ$ to $32.4^\circ$ and travel error from 1.01 to 0.68 compared
with text-only control, outperforming both training-free baselines.
It also complements learned trajectory conditioning, reducing mean
direction error from
$70.0^\circ$ to $37.3^\circ$ on SANA-WM and travel error from 2.12 to
1.64 on Lyra~2.0. Applied to the corresponding frozen base priors,
Layout Steering achieves lower translational RPE than camera-trained
Gen3C~\cite{gen3c}, TrajectoryCrafter~\cite{trajectorycrafter}, and
ReCamMaster~\cite{recammaster} without additional camera control training.
Gen3C and ReCamMaster retain lower direction and net rotation errors,
but steering achieves lower rotational RPE.

In addition to improving camera control, Layout Steering maintains
video quality comparable to the baselines. On Wan~2.2, it achieves the
highest motion smoothness and background consistency within its group.
Figure~\ref{fig:lay_results} shows more faithful camera motion and
plausible newly revealed content across video diffusion and world models.

\subsection{Long-Horizon Camera Tours}
\label{sec:exp_chain}

Camera tours require accurate control and consistent scene geometry and
appearance across generation windows. We evaluate DreamTour on the
static split of WorldScore~\cite{duan2025worldscore}, which contains
2,000 single-image worlds with one or three camera movements.
Table~\ref{tab:worldscore} reports camera controllability alongside 3D,
photometric, and style consistency.

\begin{table}[tb]
  \centering
  \caption{\textbf{Camera control and consistency on WorldScore~\cite{duan2025worldscore}.}
  The static split has 2,000 single-image worlds with one or three
  camera movements. Best results in each group are \textbf{bold}.}
  \label{tab:worldscore}

  \footnotesize
  \setlength{\tabcolsep}{4pt}
  \renewcommand{\arraystretch}{1.18}

  \begin{tabular}{@{}lcccc@{}}
    \toprule
    & & \multicolumn{3}{c}{\textbf{Scene Consistency} $\uparrow$} \\
    \cmidrule(l){3-5}
    \multirow[b]{-2}{*}{\textbf{Method}}
    & \multirow[b]{-2}{*}{\shortstack{\textbf{Camera}\\\textbf{Control} $\uparrow$}}
    & \textbf{3D}
    & \textbf{Photometric}
    & \textbf{Style}
    \\
    \midrule

    Wan~2.2~\cite{wan}
      & 27.46 & 87.74 & 93.46 & 96.73
      \\

    + CamTrol~\cite{camtrol}
      & 29.43 & 88.68 & 93.74 & 97.28
      \\

    + CamProbe~\cite{camprobe}
      & \textbf{33.21} & 67.71 & 72.79 & 66.73
      \\

    \rowcolor{oursbg}
    \textcolor{ourstext}{\textbf{+ Layout Steering (Ours)}}
      & 31.30
      & \textbf{89.05}
      & \textbf{94.26}
      & \textbf{97.96}
      \\

    \midrule

    SANA-WM~\cite{sanawm}
      & 45.88 & 84.91 & 89.34 & 82.25
      \\

    \rowcolor{oursbg}
    \textcolor{ourstext}{\textbf{+ Layout Steering (Ours)}}
      & \textbf{54.56}
      & \textbf{85.65}
      & \textbf{89.93}
      & \textbf{87.98}
      \\

    \midrule

    Lyra~2.0~\cite{lyra2026}
      & 48.83 & 84.10 & 72.88 & 60.38%
      \\

    \rowcolor{oursbg}
    \textcolor{ourstext}{\textbf{+ Layout Steering (Ours)}}
      & \textbf{49.07} & \textbf{86.39} & \textbf{75.19} & \textbf{62.35}
      \\

    \bottomrule
  \end{tabular}%
\end{table}

\begin{figure*}[t]
  \centering
  \includegraphics[width=\textwidth]{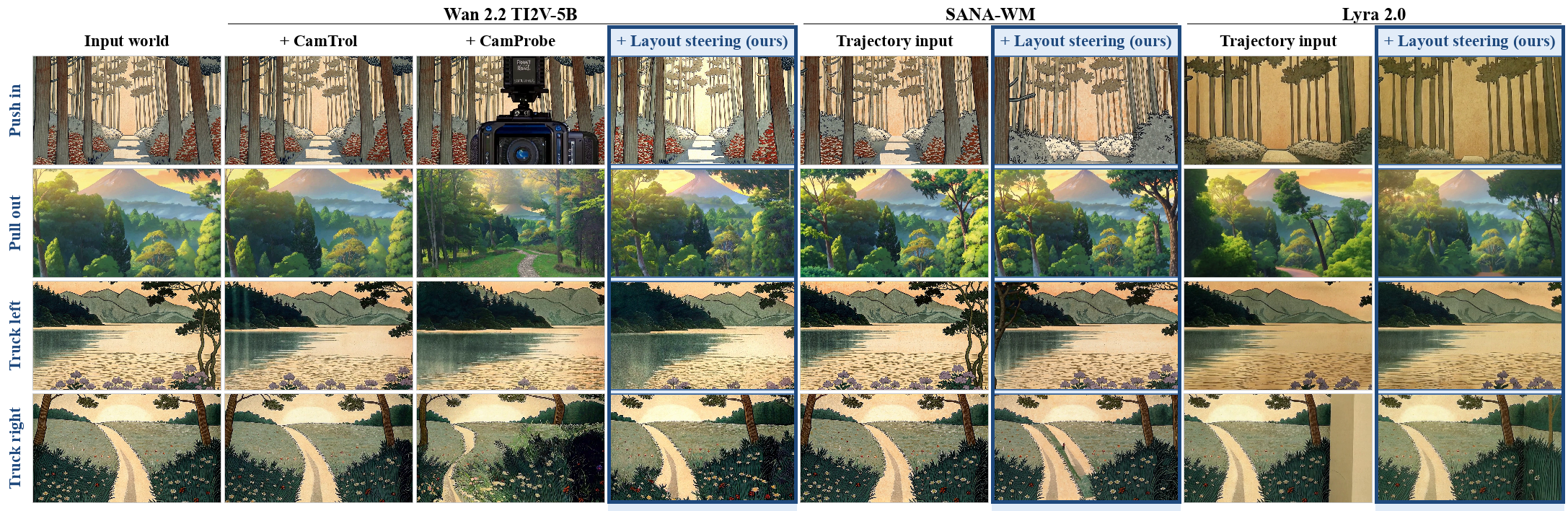}
  \caption{\textbf{Camera control and scene consistency on WorldScore~\cite{duan2025worldscore}.}
  Layout Steering yields faithful camera motion with more coherent scene
  geometry and appearance than CamTrol~\cite{camtrol} and CamProbe~\cite{camprobe}
  on Wan~2.2~\cite{wan} and trajectory conditioning alone~\cite{sanawm,lyra2026}.}
  \label{fig:ws_viz}
\end{figure*}

Layout Steering improves camera control and scene consistency across
all three models. On Wan~2.2, it outperforms the base model and CamTrol
in camera controllability while achieving the highest consistency
scores within its group. CamProbe attains higher camera controllability
but substantially lower scene consistency. Steering also complements
learned trajectory conditioning, raising SANA-WM's camera
controllability by 8.7 points and style consistency by 5.7 points.
All four WorldScore metrics improve on Lyra~2.0.
Figure~\ref{fig:ws_viz} shows more faithful camera
motion with coherent scene geometry and appearance.

\subsection{Human Evaluation}
\label{sec:human}

To complement the camera control metrics, we evaluate whether viewers
prefer videos generated with Layout Steering. Eight participants ranked
six videos for each of ten
scenes by adherence to the requested motion, from 1 for best to 6 for
worst. The videos were generated by Wan~2.2, SANA-WM, and Lyra~2.0,
each with and without Layout Steering, using the same starting image and requested motion within each scene. Clip order is randomized separately for each participant and scene. Across 80 six-way rankings, Table~\ref{tab:humanstudy} reports
mean rank and the percentage of first-place rankings.

\begin{table}[tb]
  \centering
  \caption{\textbf{Human evaluation of camera following.}
  Eight participants ranked six videos for each of ten scenes, from
  1 for best to 6 for worst. First Place is the percentage ranked first.}
  \label{tab:humanstudy}

  \footnotesize
  \setlength{\tabcolsep}{6pt}
  \renewcommand{\arraystretch}{1.18}

  \begin{tabular}{@{}cllc@{}}
    \toprule
    \shortstack{\textbf{Mean}\\\textbf{Rank} $\downarrow$}
    & \textbf{Model} & \textbf{Control Method}
    & \shortstack{\textbf{First}\\\textbf{Place} $\uparrow$} \\
    \midrule

    \rowcolor{oursbg}
    \textbf{2.84} & SANA-WM~\cite{sanawm} &
    \textcolor{ourstext}{\textbf{+ Layout Steering (Ours)}}
      & \textbf{30.00\%} \\
    \rowcolor{oursbg}
    3.03 & Wan~2.2~\cite{wan} &
    \textcolor{ourstext}{\textbf{+ Layout Steering (Ours)}}
      & 20.00\% \\
    3.20 & Lyra~2.0~\cite{lyra2026} & Trajectory Alone
      & 20.00\% \\
    \rowcolor{oursbg}
    3.58 & Lyra~2.0~\cite{lyra2026} &
    \textcolor{ourstext}{\textbf{+ Layout Steering (Ours)}}
      & 11.25\% \\
    4.00 & SANA-WM~\cite{sanawm} & Trajectory Alone
      & 11.25\% \\
    4.36 & Wan~2.2~\cite{wan} & Camera Text Alone
      & 7.50\% \\
    \bottomrule
  \end{tabular}
  \label{table:user_study}
\end{table}

\begin{table*}[t]
  \centering
  \caption{\textbf{Ablation of Layout Steering on 24 synthetic FLUX.3 scenes~\cite{flux3}.}
  We evaluate Wan~2.2~\cite{wan} on eight camera motions per scene
  following the protocol in Table~\ref{tab:camprobe}.
  Moderate, decaying low-frequency guidance during the first 40\% of
  denoising provides the best balance between camera accuracy and visual
  fidelity. Full-frequency steering yields the lowest motion errors but
  introduces visible artifacts. Alternative guides and schedules
  compromise control or background consistency.
  Shaded rows mark DreamTour's selected configuration.}
  \label{tab:ablation}

  \scriptsize
  \setlength{\tabcolsep}{2.5pt}
  \renewcommand{\arraystretch}{1.08}
  \providecolor{oursbg}{RGB}{237, 240, 245}
  \providecolor{ourstext}{RGB}{31, 73, 125}

  \begin{tabular}{@{}llcc ccc ccc@{}}
    \toprule
    & & \multicolumn{2}{c}{\textbf{Camera Control} $\downarrow$}
      & \multicolumn{3}{c}{\textbf{Camera Aim \& Displacement} $\downarrow$}
      & \multicolumn{3}{c}{\textbf{Video Quality} $\uparrow$} \\
    \cmidrule(lr){3-4}\cmidrule(lr){5-7}\cmidrule(l){8-10}
    \multirow[b]{-2}{*}{\textbf{Component}}
    & \multirow[b]{-2}{*}{\textbf{Setting}}
    & \textbf{RPE-T (m)} & \textbf{RPE-R ($^\circ$)}
    & \textbf{Dir.\ ($^\circ$)} & \textbf{Travel Err.} & \textbf{Rot. Err.}
    & \textbf{Imaging} & \textbf{Smoothness} & \textbf{Consistency} \\
    \midrule
    Baseline & Camera Text Alone & 0.066 & 1.35 & 94.0 & 1.01 & 1.06 & 70.65 & 98.06 & 96.32 \\%
    \midrule
    & Weak (0.5) & 0.049 & 0.76 & 69.8 & 0.99 & 0.88 & 70.27 & 99.08 & 97.96 \\%
    \oursrow\cellcolor{white} & \oursname{Moderate (1.0)} & 0.047 & 0.71 & 52.6 & 0.88 & 0.66 & 70.08 & 99.03 & 97.88 \\%
    \multirow{-3}{*}{Strength ($w$)} & Strong (1.5) & 0.046 & 0.73 & 49.4 & 0.86 & 0.63 & 69.69 & 98.94 & 97.34 \\%
    \midrule
    & 30\% & 0.050 & 0.69 & 56.3 & 0.91 & 0.71 & 70.15 & 99.04 & 97.96 \\%
    \oursrow\cellcolor{white} & \oursname{40\%} & 0.047 & 0.71 & 52.6 & 0.88 & 0.66 & 70.08 & 99.03 & 97.88 \\%
    \multirow{-3}{*}{Duration (K)} & 50\% & 0.046 & 0.71 & 56.1 & 0.95 & 0.66 & 70.12 & 99.08 & 97.93 \\%
    \midrule
    & Constant & 0.045 & 0.68 & 53.3 & 0.92 & 0.64 & 69.88 & 99.07 & 97.80 \\%
    \oursrow\cellcolor{white} & \oursname{Decaying} & 0.047 & 0.71 & 52.6 & 0.88 & 0.66 & 70.08 & 99.03 & 97.88 \\%
    & Full Frequency & 0.015 & 0.14 & 5.2 & 0.07 & 0.03 & 70.12 & 99.02 & 97.11 \\%
    \multirow{-4}{*}{\shortstack[l]{Schedule\ \&\\Frequency}} & ILVR-Style (Every Step, Full Weight)~\cite{ilvr} & 0.041 & 0.63 & 35.6 & 0.67 & 0.56 & 68.59 & 98.97 & 96.04 \\%
    \midrule
    \oursrow\cellcolor{white}
    & \oursname{FLUX.1 Fill (Ours)}~\cite{fluxfill} & 0.047 & 0.71 & 52.6 & 0.88 & 0.66 & 70.08 & 99.03 & 97.88 \\%
    & CamTrol~\cite{camtrol} & 0.049 & 0.80 & 65.7 & 0.98 & 0.88 & 69.42 & 99.07 & 97.71 \\%
    \multirow{-3}{*}{Guide} & CamProbe~\cite{camprobe} & 0.043 & 0.68 & 53.7 & 0.92 & 0.66 & 69.93 & 99.02 & 97.81 \\%
    \bottomrule
  \end{tabular}
\end{table*}

The participants preferred Layout Steering on Wan~2.2 and SANA-WM over
their unsteered baselines. Steering improved Lyra~2.0's measured camera
control, but participants preferred trajectory conditioning alone.
Overall, steered SANA-WM and Wan~2.2 ranked first and second, with
SANA-WM receiving 30\% of first-place votes.

\subsection{Ablation Studies}
\label{sec:ablation}

\begin{figure*}[t]
  \centering
  \input{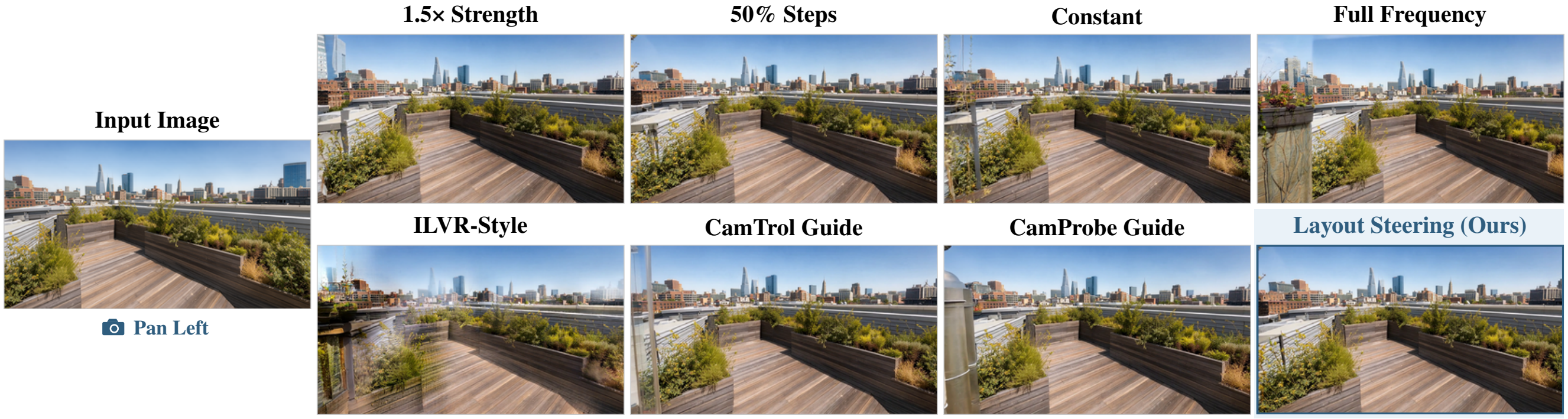}
  \caption{\textbf{Qualitative ablation of Layout Steering.}
  DreamTour coherently extends the skyline and deck during a leftward pan.
  Stronger or longer guidance introduces
  artificial structures and ghosting. Full-frequency steering and
  ILVR-style guidance~\cite{ilvr} stretch newly revealed regions, while
  guides from CamTrol~\cite{camtrol} and CamProbe~\cite{camprobe}
  distort scene geometry.
  Variants correspond to Table~\ref{tab:ablation}.}

  \label{fig:ablation_vis}
\end{figure*}

We ablate DreamTour's steering and guide construction on Wan~2.2 to
identify a configuration that balances camera accuracy with visual
fidelity. We vary one component at a time on 24 synthetic FLUX.3 scenes,
evaluating eight camera motions per scene. The default uses
expanded-canvas guidance and decaying low-frequency corrections with
strength $1.0$ during the first $40\%$ of denoising.
Table~\ref{tab:ablation} reports camera control and video quality,
while Figure~\ref{fig:ablation_vis} compares content revealed by a
leftward pan.

\textbf{Steering strength and duration.}
Weak steering leaves larger direction and travel errors. Raising the
strength from $1.0$ to $1.5$ slightly improves translational RPE
(0.047 to 0.046\,m), directional error ($52.6^\circ$ to $49.4^\circ$),
and net rotation error (0.66 to 0.63), but
reduces background consistency (97.88 to 97.34).
Figure~\ref{fig:ablation_vis} shows structural distortions under
stronger guidance. Steering for $30\%$ to $50\%$ of denoising
yields similar camera accuracy, with $40\%$ achieving the lowest
travel error (0.88), but longer guidance can introduce
ghosting and duplicated structures. These tradeoffs favor
moderate steering confined to early denoising.

\textbf{Schedule and frequency.}
Constant early guidance yields slightly higher travel error
(0.92 vs.\ 0.88) and reduces background consistency
(97.88 to 97.80).
ILVR-style guidance~\cite{ilvr} applies
full-weight low-frequency corrections at every denoising step,
degrading consistency further. Full-frequency steering achieves the lowest pose
and travel errors but reduces motion smoothness and background
consistency. Its structural artifacts are most pronounced under motions
with large disocclusions. Decaying low-frequency corrections guide
camera motion while allowing the prior to synthesize coherent,
detailed videos.

\textbf{Guide construction.}
We hold steering fixed and compare the expanded-canvas guide with
guides from CamTrol and CamProbe. CamTrol increases translational
RPE (0.047 to 0.049\,m) and net-motion errors, whereas CamProbe
slightly reduces translational RPE (0.047 to 0.043\,m) but
increases travel error (0.88 to 0.92), with unchanged rotation error.
CamTrol also introduces
structural artifacts and reduces background consistency, whereas
CamProbe preserves appearance better. The FLUX.1 Fill canvas provides
a coherent reference beyond the input field of view, supporting
camera adherence and plausible newly revealed content.

\textbf{Quality of newly revealed regions.}
Full-frame quality scores can obscure artifacts in regions exposed by
camera motion. We therefore evaluate MUSIQ~\cite{musiq} and
CLIP-IQA~\cite{wang2023exploring} on the newly revealed image bands.
Decaying low-frequency steering achieves the highest scores on both
metrics in Table~\ref{tab:ablation_percep}, outperforming constant
guidance and alternative guides. Full-frequency steering achieves
lower motion errors but weaker perceptual scores, while ILVR-style
guidance scores lowest. Figure~\ref{fig:ablation_vis} shows the
corresponding distortions in newly revealed content.

\begin{table}[t]
  \centering
  \caption{\textbf{Perceptual quality of newly revealed content.}
  We evaluate MUSIQ~\cite{musiq} and CLIP-IQA~\cite{wang2023exploring}
  on newly revealed image bands from the scenes in Table~\ref{tab:ablation}.
  Decaying low-frequency steering scores highest on both metrics.}
  \label{tab:ablation_percep}
  \footnotesize
  \setlength{\tabcolsep}{5pt}
  \renewcommand{\arraystretch}{1.18}
  \begin{tabular}{@{}lcc@{}}
    \toprule
    \textbf{Setting} & \textbf{MUSIQ} $\uparrow$ & \textbf{CLIP-IQA} $\uparrow$ \\
    \midrule
    \oursrow \oursname{Decaying (Ours)} & \textbf{63.6} & \textbf{0.597} \\
    Constant & 63.4 & 0.584 \\%
    Full Frequency & 62.6 & 0.560 \\%
    ILVR-Style (Every Step, Full Weight)~\cite{ilvr} & 60.2 & 0.525 \\%
    CamTrol Guide~\cite{camtrol} & 63.0 & 0.584 \\%
    CamProbe Guide~\cite{camprobe} & 62.8 & 0.577 \\%
    \bottomrule
  \end{tabular}
\end{table}

\section{Limitations and Future Work}
\label{sec:limitations}

DreamTour separates scene guidance from video synthesis, enabling
spatial control without retraining the video prior. Fine detail still
depends on that prior, and chained generation can accumulate
appearance drift. Persistent visual memory could complement canvas
re-anchoring to preserve appearance over longer tours. Single-image
scene completion also leaves unseen geometry ambiguous. Multi-view
evidence and uncertainty-aware guidance could improve spatial
consistency under large viewpoint changes and revisits, while pose
feedback could strengthen camera direction and travel control.

\section{Conclusion}
\label{sec:conclusion}

We presented \textbf{DreamTour}, a training-free framework that
improves camera direction and travel control in long-horizon tours
from a single photograph. A frozen inpainting prior expands the
photograph into a scene canvas, which is rendered along the requested
camera path to form a guide video. \textbf{Layout Steering} aligns
early video latents with the guide's layout through decaying
low-frequency corrections, leaving the frozen video prior to refine
appearance and dynamics. Canvas re-anchoring keeps guidance aligned
with each generated handoff view across windows. Steering improves
directional accuracy and travel calibration on VBench while
maintaining competitive video quality. It also strengthens camera
control and scene consistency across chained movements on
WorldScore. Together, these results
demonstrate that DreamTour enables video diffusion and world
models to generate long-horizon camera tours with more accurate
control and greater scene consistency.

\section*{Acknowledgements}
This work was completed during an internship at Amazon. We thank Amin Kheradmand, Avinash Ravichandran, David Stafford, Pranav Sankar and Shifan Zhu for helpful discussions and feedback.
{\small
\bibliographystyle{ieeenat_fullname}
\bibliography{main}
}
\clearpage
\maketitlesupplementary
\appendix
\raggedbottom

\providecolor{oursbg}{RGB}{237,240,245}
\providecolor{ourstext}{RGB}{31,73,125}

\setcounter{topnumber}{4}
\setcounter{bottomnumber}{2}
\setcounter{totalnumber}{6}
\setcounter{dbltopnumber}{4}
\renewcommand{\topfraction}{0.90}
\renewcommand{\bottomfraction}{0.75}
\renewcommand{\textfraction}{0.08}
\renewcommand{\floatpagefraction}{0.55}
\renewcommand{\dbltopfraction}{0.92}
\renewcommand{\dblfloatpagefraction}{0.72}
\setlength{\textfloatsep}{10pt plus 2pt minus 2pt}
\setlength{\dbltextfloatsep}{10pt plus 2pt minus 2pt}
\setlength{\floatsep}{8pt plus 2pt minus 2pt}
\setlength{\dblfloatsep}{8pt plus 2pt minus 2pt}
\makeatletter
\setlength{\@fptop}{0pt}
\setlength{\@dblfptop}{0pt}
\setlength{\@fpsep}{12pt plus 2pt}
\setlength{\@dblfpsep}{12pt plus 2pt}
\setlength{\@fpbot}{0pt plus 1fil}
\setlength{\@dblfpbot}{0pt plus 1fil}
\makeatother

\section{Layout Steering as Diffusion Guidance}
\label{app:guidance}

Layout Steering is a sampling-time guidance mechanism that aligns
intermediate video latents with the coarse spatial layout of a
trajectory-rendered guide. The guide is constructed from an expanded
input image and estimated scene depth. During early denoising, a
filtered latent correction encourages the frozen video prior to follow
the requested camera direction and travel distance while leaving
appearance and dynamics to the prior.

\noindent \textbf{Connection to diffusion guidance.}
Classifier guidance~\cite{dhariwal2021diffusion} adjusts denoising
using $\nabla_z\log p(y\mid z)$ from a classifier trained on noisy
samples. Classifier-free guidance~\cite{ho2022cfg} combines
conditional and unconditional predictions as
$\hat\epsilon=(1+s)\epsilon_{\mathrm{cond}}-
s\epsilon_{\mathrm{uncond}}$. Unlike these learned guidance signals,
Layout Steering directly measures the spatial mismatch between the
video latent and the rendered guide, without training an auxiliary
model or requiring an extra video-model evaluation.

Following the latent update in Section~\ref{sec:steering}, let $\mathcal{L}$ be a
spatial Gaussian low-pass filter and $\mathcal{P}_V$ the
model-dependent normalization. When
$A=\mathcal{P}_V\mathcal{L}$ is symmetric and positive semidefinite,
the correction is a gradient step on the quadratic potential
\begin{equation}
\begin{aligned}
E_i(z) &= \tfrac{1}{2}\big\langle z-\tilde z^\Omega_{i+1},\,
A(z-\tilde z^\Omega_{i+1})\big\rangle,\\
z_{i+1} &= z'_{i+1}-w_i\nabla_z E_i(z'_{i+1}).
\end{aligned}
\label{eq:guidance_potential}
\end{equation}
Here $z'_{i+1}$ is the video latent after an ordinary denoising
step and $\tilde z^\Omega_{i+1}$ is the encoded guide noised to the
same level. Under this assumption,
$\nabla_z E_i(z)=A(z-\tilde z^\Omega_{i+1})$, recovering the
explicit filtered correction used by Layout Steering.
The equivalence is exact for an appropriate self-adjoint filter
and normalization and approximate when boundary handling breaks
that assumption. The implementation applies the filtered
correction directly rather than evaluating the potential.

\noindent \textbf{Frequency, normalization, and duration.}
The low-pass filter operates independently on each latent frame and
channel, with a Gaussian standard deviation of three latent cells.
It attenuates high-frequency discrepancies in the approximate guide,
reducing the transfer of reprojection artifacts while steering the
coarse structure that controls camera direction and travel.
For Wan, CogVideoX, HunyuanVideo, and Cosmos, $\mathcal{P}_V$
removes the spatial mean of the correction for each latent frame
and channel to limit shifts in color and exposure. For SANA-Video
and SANA-WM, the full correction is retained because their
compressed latent statistics also carry spatial layout.

Guidance is restricted to the first $K$ of $N$ denoising steps,
with the linear schedule $w_i=w[1-(i+1)/K]$ for
$i=0,\ldots,K-1$. The remaining $N-K$ steps use the frozen sampler
without additional steering. This concentrates guidance on early
layout formation and preserves the prior's ability to refine
appearance and dynamics.

\noindent \textbf{Guidance strength and duration.}
The ablations in Table~\ref{tab:ablation} of the main paper vary correction strength,
duration, frequency content, and guide construction. Stronger
steering can reduce direction and travel errors but may introduce
structures copied from inaccurate guide regions. A constant weight
or a longer guidance interval can also introduce ghosting and
reduce background consistency. Full-frequency correction yields
particularly low camera errors but more visible artifacts.
The evaluated ILVR-style variant~\cite{ilvr} applies full-weight
low-frequency corrections throughout denoising and similarly
reduces visual quality. These observations motivate moderate,
decaying low-frequency steering, consistent with the coarse-to-fine
behavior of diffusion models~\cite{ac3d} and interval-limited
guidance~\cite{kynkaanniemi2024applying}.

\begin{figure}[!htbp]
  \centering
  \includegraphics[width=\linewidth,height=0.27\textheight,keepaspectratio]{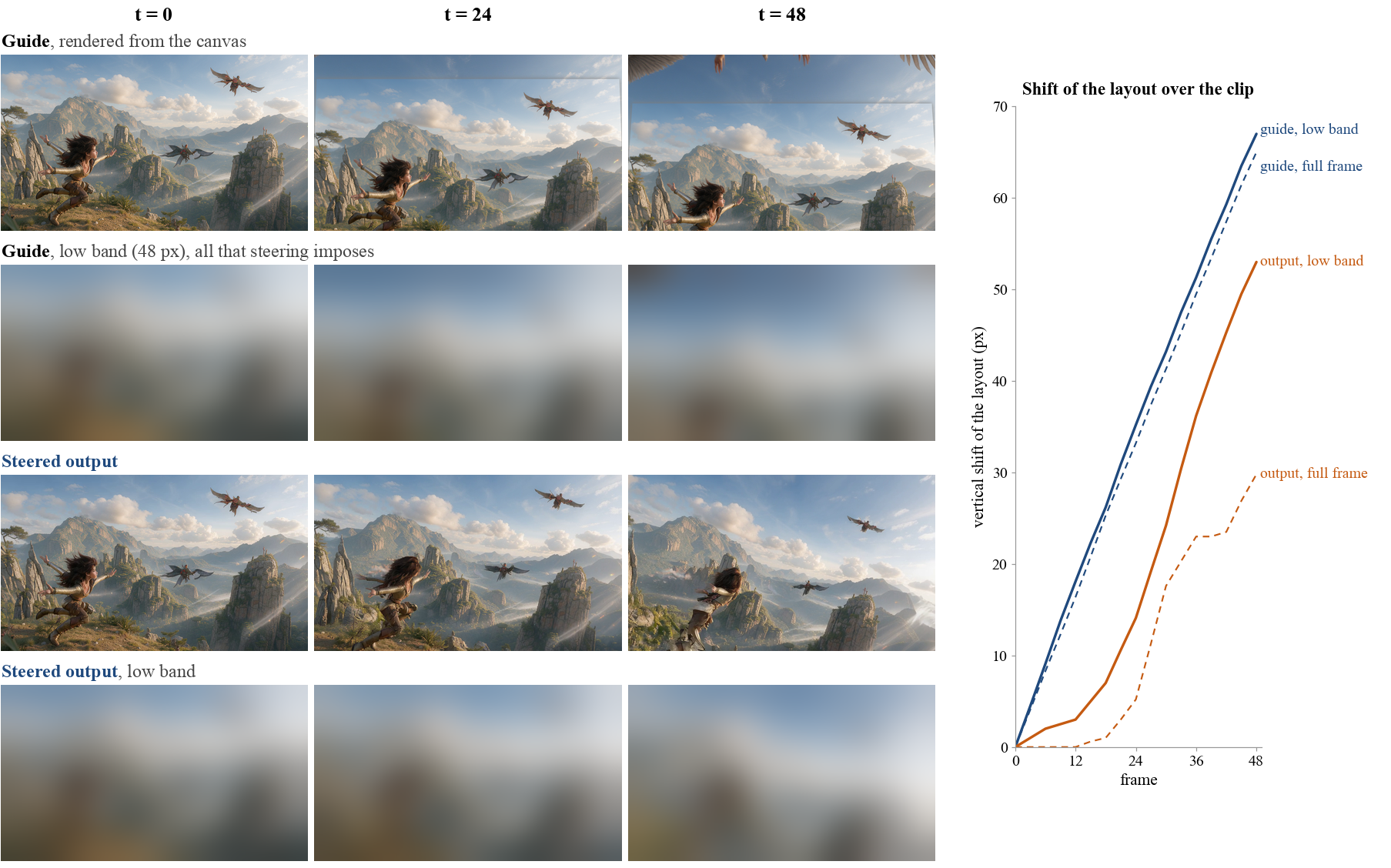}
  \caption{\textbf{Low-frequency layout during an upward camera tilt.}
  On Wan~2.2 5B, Layout Steering follows the guide's coarse
  spatial displacement without directly copying its high-frequency
  appearance or reprojection artifacts.}
  \label{fig:lowfreq_tilt}
\end{figure}

\subsection{Position-Level Steering}
\label{app:position}

We investigate two additional mechanisms for expressing the
requested camera displacement in noise initialization or token
positions. Both operate without modifying the pretrained weights
and are evaluated together with Layout Steering.

\noindent \textbf{Noise transport.}
Instead of independent initial noise for each frame, we draw one
Gaussian field $\mathcal{E}$ over the expanded scene at latent
resolution, with spatial dimensions
$(H'+\lceil m_y/8\rceil)\times(W'+\lceil m_x/8\rceil)$.
For frame $j$, a crop shifted by the requested image-space offset
$o_{t_j}$ defines its initial noise
\begin{equation}
\epsilon_j=\operatorname{Crop}(\mathcal{E},o_{t_j}/8).
\label{eq:noise_transport}
\end{equation}
Corresponding scene regions therefore reuse the same noise across
frames. Unlike Go-with-the-Flow~\cite{gowiththeflow}, this variant
does not fine-tune the video prior to recognize transported noise.

\noindent \textbf{Rotary steering.}
Wan~2.2 uses rotary embeddings for token positions along time,
height, and width. For frame $j$, we shift horizontal or vertical
positions by $g\,o_{t_j}/16$ tokens according to the requested
camera travel. For forward travel, token positions are contracted
toward the image centre using the frame's zoom factor. These
changes encourage attention to associate corresponding scene
regions across frames. We set $g=0.5$ because larger shifts tend
to overshoot the requested travel, and apply the shift to both
Wan~2.2 experts.

\noindent \textbf{Results.}
Table~\ref{tab:steering} evaluates the two variants on the same
24 synthetic FLUX.3 scenes and eight camera motions as Table~\ref{tab:ablation}
of the main paper. Neither variant improves the overall balance
of camera control and video quality over Layout Steering alone.

\begin{table*}[!tp]
  \centering
  \caption{\textbf{Position-level steering on Wan~2.2 5B.}
  Noise transport and rotary positional shifts are added to
  Layout Steering on 24 synthetic FLUX.3 scenes with eight camera
  motions per scene. Lower is better for camera errors and higher
  is better for video-quality metrics.}
  \label{tab:steering}
  \scriptsize
  \setlength{\tabcolsep}{3.5pt}
  \renewcommand{\arraystretch}{1.16}
  \begin{adjustbox}{max width=\textwidth}
  \begin{tabular}{@{}lcccccccc@{}}
    \toprule
    & \multicolumn{2}{c}{\textbf{Camera Control} $\downarrow$}
    & \multicolumn{3}{c}{\textbf{Camera Aim \& Displacement} $\downarrow$}
    & \multicolumn{3}{c}{\textbf{Video Quality} $\uparrow$} \\
    \cmidrule(lr){2-3}\cmidrule(lr){4-6}\cmidrule(l){7-9}
    \textbf{Setting} & \textbf{RPE-T (m)} & \textbf{RPE-R ($^\circ$)}
    & \textbf{Dir. ($^\circ$)} & \textbf{Travel Err.} & \textbf{Rot. Err.}
    & \textbf{Imaging} & \textbf{Smoothness} & \textbf{Consistency} \\
    \midrule
    \rowcolor{oursbg}
    \textcolor{ourstext}{\textbf{Layout Steering}}
      & 0.047 & 0.71 & 52.6 & 0.88 & 0.66 & 70.08 & 99.03 & 97.88 \\
    + Noise transport
      & 0.049 & 0.76 & 68.3 & 0.97 & 0.92 & 70.19 & 98.79 & 97.94 \\
    + Rotary shift
      & 0.048 & 0.78 & 56.6 & 0.96 & 0.62 & 68.92 & 98.63 & 96.83 \\
    + Noise transport + rotary shift
      & 0.054 & 0.75 & 53.1 & 0.95 & 0.54 & 68.93 & 98.31 & 96.76 \\
    \bottomrule
  \end{tabular}
  \end{adjustbox}
\end{table*}

\section{Additional Ablations}
\label{app:ablations}

We extend the Wan~2.2 ablations in Table~\ref{tab:ablation} of the main paper to
the trajectory-conditioned world models SANA-WM and Lyra~2.0.
Each model retains its native trajectory input. All ablations use
24 synthetic FLUX.3 scenes and eight requested camera motions per
scene, rather than the 93-prompt VBench evaluation in Section~\ref{sec:exp_wm}.
We additionally examine guide-depth estimation and the contribution
of image expansion to guide construction.

\noindent \textbf{World-model steering.}
Tables~\ref{tab:ablation_sanawm} and~\ref{tab:ablation_lyra}
report model-specific strength, duration, and frequency choices.
For SANA-WM, weight 1.5 with linear decay over the first 50\% of
denoising gives a favorable balance of travel accuracy and
appearance. Lyra~2.0 uses weight 1.0 over the first 40\% of
steps. Stronger or more persistent correction does not improve
all metrics simultaneously, reinforcing the need to tune each
frozen prior separately.

\begin{table*}[!tp]
  \centering
  \caption{\textbf{Layout Steering ablation on SANA-WM.} Evaluated on 24 synthetic FLUX.3 scenes with eight camera motions per scene. The model retains its native trajectory conditioning in every setting. The shaded rows mark the selected configuration, with weight 1.5 applied over the first 50\% of denoising with linear decay. Camera errors are lower-is-better, while quality scores are higher-is-better.}
  \label{tab:ablation_sanawm}
  \scriptsize
  \setlength{\tabcolsep}{3pt}
  \renewcommand{\arraystretch}{1.05}
  \begin{adjustbox}{max width=\textwidth}
  \begin{tabular*}{\textwidth}{@{\extracolsep{\fill}}llccccc@{}}
    \toprule
    \multicolumn{2}{@{}l}{\textbf{Camera metrics}} & \multicolumn{5}{c@{}}{\textbf{Lower is better $\downarrow$}} \\
    \midrule
    \textbf{Component} & \textbf{Setting} & \textbf{RPE-T (m)} & \textbf{RPE-R ($^\circ$)} & \textbf{Dir. ($^\circ$)} & \textbf{Travel Err.} & \textbf{Rot. Err.} \\
    \midrule
    Baseline & Trajectory Alone & 0.046 & 0.56 & 65.7 & 0.97 & 1.01 \\
    \addlinespace[2pt]
    Strength & Weak (1.0) & 0.037 & 0.36 & 30.9 & 0.72 & 0.47 \\
    \rowcolor{oursbg}
     & \textcolor{ourstext}{\textbf{Moderate (1.5)}} & 0.034 & 0.33 & 24.4 & 0.67 & 0.46 \\
     & Strong (2.0) & 0.036 & 0.36 & 27.7 & 0.73 & 0.45 \\
    \addlinespace[2pt]
    Duration & 30\% & 0.035 & 0.37 & 35.8 & 0.72 & 0.46 \\
    \rowcolor{oursbg}
     & \textcolor{ourstext}{\textbf{50\%}} & 0.034 & 0.33 & 24.4 & 0.67 & 0.46 \\
     & 60\% & 0.035 & 0.34 & 26.3 & 0.69 & 0.47 \\
    \addlinespace[2pt]
    Schedule / frequency & Constant & 0.032 & 0.33 & 23.3 & 0.66 & 0.51 \\
    \rowcolor{oursbg}
     & \textcolor{ourstext}{\textbf{Decaying}} & 0.034 & 0.33 & 24.4 & 0.67 & 0.46 \\
     & Full Frequency & 0.046 & 0.50 & 55.4 & 0.97 & 0.77 \\
     & Every Step, Full Weight & 0.035 & 0.34 & 21.7 & 0.64 & 0.49 \\
    \bottomrule
  \end{tabular*}
  \end{adjustbox}

  \vspace{5pt}

  \begin{adjustbox}{max width=\textwidth}
  \begin{tabular*}{\textwidth}{@{\extracolsep{\fill}}llccccc@{}}
    \toprule
    \multicolumn{2}{@{}l}{\textbf{Appearance and newly revealed content}} & \multicolumn{5}{c@{}}{\textbf{Higher is better $\uparrow$}} \\
    \midrule
    \textbf{Component} & \textbf{Setting} & \textbf{Imaging} & \textbf{Smoothness} & \textbf{Consistency} & \textbf{MUSIQ} & \textbf{CLIP-IQA} \\
    \midrule
    Baseline & Trajectory Alone & 71.62 & 99.26 & 97.79 & 65.9 & 0.651 \\
    \addlinespace[2pt]
    Strength & Weak (1.0) & 71.46 & 99.43 & 97.73 & 64.6 & 0.653 \\
    \rowcolor{oursbg}
     & \textcolor{ourstext}{\textbf{Moderate (1.5)}} & 71.46 & 99.42 & 97.46 & 64.3 & 0.647 \\
     & Strong (2.0) & 71.38 & 99.39 & 97.28 & 63.3 & 0.601 \\
    \addlinespace[2pt]
    Duration & 30\% & 71.64 & 99.42 & 97.64 & 64.5 & 0.640 \\
    \rowcolor{oursbg}
     & \textcolor{ourstext}{\textbf{50\%}} & 71.46 & 99.42 & 97.46 & 64.3 & 0.647 \\
     & 60\% & 71.45 & 99.43 & 97.63 & 64.8 & 0.653 \\
    \addlinespace[2pt]
    Schedule / frequency & Constant & 71.41 & 99.35 & 97.51 & 65.1 & 0.653 \\
    \rowcolor{oursbg}
     & \textcolor{ourstext}{\textbf{Decaying}} & 71.46 & 99.42 & 97.46 & 64.3 & 0.647 \\
     & Full Frequency & 65.48 & 99.40 & 95.97 & 57.3 & 0.572 \\
     & Every Step, Full Weight & 71.48 & 99.42 & 97.52 & 65.6 & 0.649 \\
    \bottomrule
  \end{tabular*}
  \end{adjustbox}
\end{table*}

\begin{table*}[!tp]
  \centering
  \caption{\textbf{Layout Steering ablation on Lyra~2.0.} Evaluated on 24 synthetic FLUX.3 scenes with eight camera motions per scene. The model retains its native trajectory conditioning in every setting. The shaded rows mark the selected configuration, with weight 1.0 applied over the first 40\% of denoising with linear decay. Camera errors are lower-is-better, while quality scores are higher-is-better.}
  \label{tab:ablation_lyra}
  \scriptsize
  \setlength{\tabcolsep}{3pt}
  \renewcommand{\arraystretch}{1.05}
  \begin{adjustbox}{max width=\textwidth}
  \begin{tabular*}{\textwidth}{@{\extracolsep{\fill}}llccccc@{}}
    \toprule
    \multicolumn{2}{@{}l}{\textbf{Camera metrics}} & \multicolumn{5}{c@{}}{\textbf{Lower is better $\downarrow$}} \\
    \midrule
    \textbf{Component} & \textbf{Setting} & \textbf{RPE-T (m)} & \textbf{RPE-R ($^\circ$)} & \textbf{Dir. ($^\circ$)} & \textbf{Travel Err.} & \textbf{Rot. Err.} \\
    \midrule
    Baseline & Trajectory Alone & 0.066 & 0.64 & 54.5 & 1.59 & 0.82 \\
    \addlinespace[2pt]
    Strength & Weak (0.5) & 0.064 & 0.62 & 52.4 & 1.18 & 0.78 \\
    \rowcolor{oursbg}
     & \textcolor{ourstext}{\textbf{Moderate (1.0)}} & 0.048 & 0.57 & 51.5 & 1.01 & 0.74 \\
     & Strong (1.5) & 0.053 & 0.62 & 51.5 & 1.20 & 0.84 \\
    \addlinespace[2pt]
    Duration & 30\% & 0.054 & 0.62 & 52.6 & 1.28 & 0.84 \\
    \rowcolor{oursbg}
     & \textcolor{ourstext}{\textbf{40\%}} & 0.048 & 0.57 & 51.5 & 1.01 & 0.74 \\
     & 50\% & 0.049 & 0.61 & 51.3 & 1.16 & 0.82 \\
    \addlinespace[2pt]
    Schedule / frequency & Constant & 0.046 & 0.56 & 48.0 & 0.94 & 0.66 \\
    \rowcolor{oursbg}
     & \textcolor{ourstext}{\textbf{Decaying}} & 0.048 & 0.57 & 51.5 & 1.01 & 0.74 \\
     & Full Frequency & 0.050 & 0.61 & 47.9 & 1.29 & 0.72 \\
     & Every Step, Full Weight & 0.040 & 0.47 & 39.2 & 0.63 & 0.48 \\
    \bottomrule
  \end{tabular*}
  \end{adjustbox}

  \vspace{5pt}

  \begin{adjustbox}{max width=\textwidth}
  \begin{tabular*}{\textwidth}{@{\extracolsep{\fill}}llccccc@{}}
    \toprule
    \multicolumn{2}{@{}l}{\textbf{Appearance and newly revealed content}} & \multicolumn{5}{c@{}}{\textbf{Higher is better $\uparrow$}} \\
    \midrule
    \textbf{Component} & \textbf{Setting} & \textbf{Imaging} & \textbf{Smoothness} & \textbf{Consistency} & \textbf{MUSIQ} & \textbf{CLIP-IQA} \\
    \midrule
    Baseline & Trajectory Alone & 75.51 & 99.19 & 95.76 & 64.2 & 0.601 \\
    \addlinespace[2pt]
    Strength & Weak (0.5) & 72.33 & 99.21 & 95.73 & 64.4 & 0.619 \\
    \rowcolor{oursbg}
     & \textcolor{ourstext}{\textbf{Moderate (1.0)}} & 71.92 & 99.21 & 95.81 & 64.8 & 0.620 \\
     & Strong (1.5) & 72.21 & 99.22 & 95.67 & 64.5 & 0.618 \\
    \addlinespace[2pt]
    Duration & 30\% & 72.25 & 99.21 & 95.65 & 64.5 & 0.620 \\
    \rowcolor{oursbg}
     & \textcolor{ourstext}{\textbf{40\%}} & 71.92 & 99.21 & 95.81 & 64.8 & 0.620 \\
     & 50\% & 72.10 & 99.22 & 95.75 & 63.7 & 0.614 \\
    \addlinespace[2pt]
    Schedule / frequency & Constant & 71.86 & 99.22 & 95.66 & 64.5 & 0.613 \\
    \rowcolor{oursbg}
     & \textcolor{ourstext}{\textbf{Decaying}} & 71.92 & 99.21 & 95.81 & 64.8 & 0.620 \\
     & Full Frequency & 70.28 & 99.25 & 93.45 & 60.2 & 0.549 \\
     & Every Step, Full Weight & 71.88 & 99.12 & 94.37 & 63.5 & 0.582 \\
    \bottomrule
  \end{tabular*}
  \end{adjustbox}
\end{table*}

\noindent \textbf{Guide depth.}
DreamTour estimates depth on the expanded canvas with $\pi^3$X
and calibrates its scale using the input image's metric depth
from Depth Anything~3, as described in Section~\ref{sec:canvas}.
The ablation changes only the guide's canvas-depth estimator.
Table~\ref{tab:ablation_depth} shows the smallest direction,
travel, and rotation errors with $\pi^3$X among the tested
backends, even though the RPE-T and RPE-R differences are modest.

\begin{table*}[!tp]
  \centering
  \caption{\textbf{Guide-depth robustness on Wan~2.2 5B.}
  We replace only the depth estimator used to render the guide,
  keeping weight 1.0 with linear decay over the first 40\% of
  denoising. The evaluation uses 24 FLUX.3 scenes with eight
  camera motions per scene. A flat plane removes scene-specific
  depth variation. Lower is better for camera errors and higher
  is better for video-quality metrics.}
  \label{tab:ablation_depth}
  \scriptsize
  \setlength{\tabcolsep}{3.5pt}
  \renewcommand{\arraystretch}{1.05}
  \begin{adjustbox}{max width=\textwidth}
  \begin{tabular}{@{}lcccccccc@{}}
    \toprule
    & \multicolumn{2}{c}{\textbf{Camera Control} $\downarrow$}
    & \multicolumn{3}{c}{\textbf{Camera Aim \& Displacement} $\downarrow$}
    & \multicolumn{3}{c}{\textbf{Video Quality} $\uparrow$} \\
    \cmidrule(lr){2-3}\cmidrule(lr){4-6}\cmidrule(l){7-9}
    \textbf{Guide depth} & \textbf{RPE-T (m)} & \textbf{RPE-R ($^\circ$)}
    & \textbf{Dir. ($^\circ$)} & \textbf{Travel Err.} & \textbf{Rot. Err.}
    & \textbf{Imaging} & \textbf{Smoothness} & \textbf{Consistency} \\
    \midrule
    Flat plane & 0.051 & 0.71 & 60.4 & 0.97 & 0.78 & 70.23 & 99.05 & 97.96 \\
    Depth Anything V2 (small) & 0.048 & 0.71 & 62.3 & 0.98 & 0.79 & 70.20 & 99.06 & 97.94 \\
    \rowcolor{oursbg}
    \textcolor{ourstext}{\textbf{$\pi^3$X (guide depth)}}
      & 0.047 & 0.71 & 52.6 & 0.88 & 0.66 & 70.08 & 99.03 & 97.88 \\
    Depth Anything 3 & 0.050 & 0.71 & 64.0 & 0.98 & 0.76 & 70.25 & 99.06 & 97.95 \\
    \bottomrule
  \end{tabular}
  \end{adjustbox}
\end{table*}

\noindent \textbf{Role of canvas expansion.}
We compare guides rendered from the expanded canvas and directly
from the input image on 48 cases, comprising six starting scenes
and eight motions. Without expansion, newly exposed regions are
filled by border extension. With expansion, direction error is
$53.1^\circ$, 62\% of clips have error below $45^\circ$,
median travel and rotation ratios are 0.14 and 0.62, and 17\%
of clips are within $\pm25\%$ of requested travel. The
corresponding values without expansion are $49.4^\circ$, 65\%,
0.21, 0.66, and 25\%. VBench imaging, motion smoothness, and
background consistency are 74.0, 99.03, and 97.4 with expansion,
compared with 73.9, 99.03, and 97.3 without it.

These small-motion results do not show a consistent quantitative
advantage for expansion. The evaluated translations are around
0.58\,m and rotations around $8.3^\circ$, revealing only narrow
image bands. The expanded canvas is intended to provide coherent
content outside the input field of view, which becomes more
important for larger viewpoint changes and chained camera tours
in Section~\ref{sec:exp_chain}.

\section{Camera-Control Evaluation Protocol}
\label{app:camprobe}

\subsection{Dataset and Pose Estimation}

The single-window evaluation in Table~\ref{tab:camprobe} of the main paper follows
the eight-motion CamProbe protocol~\cite{camprobe} on 93 VBench
prompts, giving up to 744 motion-prompt cases per method. Each clip
contains 49 frames at 16\,fps. Most models are evaluated at
$832\times480$ resolution. Cosmos-Predict1 and SANA-WM produce
$1280\times704$ outputs and are evaluated at $832\times458$ to
preserve their aspect ratio. Requested motions cover translation,
rotation, and arcs, with starting scenes and trajectories shared
across compared methods.

We estimate camera poses and intrinsics directly from RGB frames
using the metric Depth Anything~3 checkpoint
\texttt{DA3NESTED-GIANT-LARGE-1.1}, with OpenCV camera
coordinates. The generated clip and trajectory-rendered guide
are reconstructed jointly in a shared coordinate frame. Metric
scale is set using only the estimated median depth of the input
image, without post-hoc trajectory alignment or scale fitting.
The requested motions follow CamProbe's motion files, including
rotations of approximately $8.3^\circ$ and translations scaled
relative to scene depth.

\noindent \textbf{Relative pose error.}
We compute RPE-T and RPE-R on unaligned estimated trajectories,
comparing relative camera motions across gaps of 2, 4, 8, and
12 frames. This reports local translation and rotation error
without a fitted Sim(3) transform. In particular, Sim(3)
alignment could mask the distance travelled by rescaling a
nearly stationary trajectory. We follow the evaluation setup of
CamProbe~\cite{camprobe} but re-estimate all compared outputs
with the same Depth Anything~3 pipeline rather than reproducing
published scores. Our RPE-R values are approximately
$0.3^\circ$ to $1.4^\circ$, whereas the original CamProbe
publication reports approximately $2.7^\circ$ to $3.2^\circ$.
We therefore compare results within the reimplementation and
do not pool RPE values from different papers.

\subsection{Implementation and Baselines}

\noindent \textbf{Model-specific steering.}
We apply a spatial Gaussian filter with standard deviation three
latent cells and restrict the correction to early denoising.
Unless otherwise specified, guidance decays linearly to zero.
The selected Wan~2.2 5B configuration uses weight 1.0 over the
first 40\% of steps, as in Table~\ref{tab:ablation} of the main paper. SANA-WM
uses weight 1.5 for 30 of 60 steps. Lyra~2.0 uses weight 1.0 over the first 40\% of denoising. The separately evaluated SANA-Video~2.0 checkpoint uses 40 of
50 steps, Cosmos-Predict1 uses 11 of 36, and Wan~2.1,
CogVideoX-Fun, and HunyuanVideo~1.5 use 15 of 50.
These settings are chosen per model on the 24-scene ablation
set and should not be interpreted as the Wan~2.2 setting.

Wan, CogVideoX, HunyuanVideo, and Cosmos use a mean-free
correction to limit changes in per-frame latent channel means.
SANA-Video and SANA-WM retain the full correction because
these statistics also encode spatial layout. All pretrained
weights remain frozen, and camera-conditioned world models
retain their native trajectory input.

The CamProbe evaluation in Table~\ref{tab:camprobe} and the 24-scene ablations
in Table~\ref{tab:ablation} measure different clip populations and therefore
need not produce identical scores. The runtime study below
uses a separate 81-frame configuration.

\noindent \textbf{Training-free baselines.}
CamTrol~\cite{camtrol} uses ZoeDepth to lift the input image
into a point cloud, renders novel views, and inpaints exposed
regions. The video sampler restarts from a noised version of
that layout video at step 20 in a 50-step setting. We also
adapt the released CamProbe latent-warping procedure to frozen
Wan~2.2 5B. During the first 20\% of denoising steps,
we decode the predicted clean latent, estimate per-frame
MiDaS-small depth, warp according to the requested camera
displacement with border padding, re-noise, and resume sampling.
The released implementation targets HunyuanVideo, so this
is an adaptation rather than an unchanged checkpoint result.

The requested camera trajectories are shared across methods.
CamTrol completes 742 of 744 motion-prompt cases, and its
reported row uses the completed clips. Hard latent replacement can copy rendering holes and stretched regions, so we retain the standard CamTrol configuration for the comparison.
Transporting initial noise without Go-with-the-Flow's
noise-adaptation training substantially degrades output
quality and is not included as a training-free baseline.

\noindent \textbf{Learned camera-control baselines.}
For Gen3C, TrajectoryCrafter, and ReCamMaster, we evaluate
released camera-trained models and compare Layout Steering
on the corresponding frozen video priors. DreamTour never
trains or modifies video-prior weights. SANA-WM and Lyra~2.0
are evaluated both with native trajectory conditioning alone
and with Layout Steering added. Cosmos-Predict1's text filter
rejects seven of 93 prompts, leaving 688 valid motion-prompt
cases for its steered evaluation rather than 744.

\begin{table*}[!tp]
  \centering
  \caption{\textbf{Camera travel under six requests.} Summed median optical flow over each approximately five-second clip, with frames resized to 512 pixels wide. Values are mean $\pm$ standard deviation over the observed clips, with $n$ pooled across repeated seeds. Positive lateral camera travel is leftward and positive vertical travel is upward. Positive zoom denotes forward travel. Followed is the percentage matching the requested direction, with stationary lateral travel defined as less than 10 pixels.}
  \label{tab:bias_full}
  \footnotesize
  \begin{adjustbox}{max width=\textwidth}
  \setlength{\tabcolsep}{5pt}
  \renewcommand{\arraystretch}{1.05}
  \begin{tabular}{@{}llrrrrr@{}}
    \toprule
    Prior & Camera sentence & $n$ & Lateral (px) & Vertical (px) & Zoom & Followed (\%) \\
    \midrule
    \multirow{6}{*}{Wan~2.2} & No sentence & 120 & $-1 \pm 8$ & $1 \pm 3$ & +0.08 & -- \\
     & Stay still & 24 & $-1 \pm 2$ & $1 \pm 2$ & +0.04 & 100 \\
     & Truck right & 120 & $-59 \pm 47$ & $1 \pm 4$ & +0.20 & 85 \\
     & Truck left & 120 & $62 \pm 51$ & $3 \pm 7$ & +0.15 & 93 \\
     & Forward & 24 & $-8 \pm 22$ & $6 \pm 10$ & +0.77 & 100 \\
     & Rise & 24 & $-2 \pm 8$ & $13 \pm 14$ & +0.38 & 83 \\
    \midrule
    \multirow{6}{*}{SANA-Video~2.0 5B} & No sentence & 120 & $30 \pm 34$ & $3 \pm 11$ & +0.77 & -- \\
     & Stay still & 24 & $19 \pm 26$ & $3 \pm 10$ & +1.10 & 58 \\
     & Truck right & 120 & $30 \pm 41$ & $4 \pm 10$ & +0.87 & 13 \\
     & Truck left & 120 & $29 \pm 39$ & $4 \pm 10$ & +0.88 & 87 \\
     & Forward & 24 & $7 \pm 17$ & $7 \pm 15$ & +1.44 & 100 \\
     & Rise & 24 & $10 \pm 14$ & $20 \pm 39$ & +0.76 & 75 \\
    \midrule
    \multirow{6}{*}{CogVideoX~1.5 5B} & No sentence & 44 & $-9 \pm 24$ & $-0 \pm 7$ & +0.15 & -- \\
     & Stay still & 24 & $-7 \pm 32$ & $-1 \pm 6$ & +0.22 & 46 \\
     & Truck right & 44 & $-14 \pm 36$ & $-1 \pm 8$ & +0.11 & 70 \\
     & Truck left & 43 & $-13 \pm 37$ & $-1 \pm 7$ & +0.11 & 33 \\
     & Forward & 24 & $-6 \pm 36$ & $0 \pm 4$ & +0.35 & 83 \\
     & Rise & 24 & $0 \pm 24$ & $2 \pm 11$ & +0.23 & 75 \\
    \midrule
    \multirow{6}{*}{SANA-WM} & No sentence & 120 & $1 \pm 2$ & $0 \pm 0$ & +0.04 & -- \\
     & Stay still & 24 & $0 \pm 2$ & $0 \pm 0$ & $-$0.00 & 100 \\
     & Truck right & 118 & $-45 \pm 35$ & $0 \pm 1$ & +0.01 & 99 \\
     & Truck left & 118 & $49 \pm 40$ & $0 \pm 1$ & $-$0.02 & 98 \\
     & Forward & 23 & $-3 \pm 4$ & $-1 \pm 2$ & +0.42 & 100 \\
     & Rise & 23 & $1 \pm 2$ & $25 \pm 16$ & +0.12 & 100 \\
    \midrule
    \multirow{6}{*}{Lyra~2.0} & No sentence & 72 & $-0 \pm 0$ & $-0 \pm 0$ & +0.00 & -- \\
     & Stay still & 24 & $-0 \pm 0$ & $-0 \pm 0$ & +0.01 & 100 \\
     & Truck right & 72 & $-142 \pm 58$ & $-0 \pm 1$ & $-$0.05 & 100 \\
     & Truck left & 72 & $146 \pm 57$ & $-0 \pm 0$ & $-$0.01 & 100 \\
     & Forward & 24 & $0 \pm 1$ & $1 \pm 1$ & +1.34 & 100 \\
     & Rise & 24 & $0 \pm 1$ & $104 \pm 50$ & +0.38 & 100 \\
    \bottomrule
  \end{tabular}
  \end{adjustbox}

\end{table*}

\begin{table*}[!tp]
  \centering
  \caption{\textbf{Scene-composition cues and unprompted camera travel.}
  The upper block reports sign agreement in percent between unprompted
  travel and each image cue, with 50\% chance agreement for binary
  cues. The requested-direction, mirror-reversal, and blur-avoidance
  rows are separate diagnostic tests. The lower block reports
  Spearman correlations between cue strength and observed travel.
  Lateral and vertical cues compare opposite sides of the image.
  Open centre compares central and peripheral inverse depth.
  Vertical and zoom correlations use only clips with appreciable
  spontaneous travel on the corresponding axis.}
  \label{tab:bias_predictors}
  \scriptsize
  \setlength{\tabcolsep}{3pt}
  \renewcommand{\arraystretch}{1.05}
  \begin{adjustbox}{max width=\textwidth}
  \begin{tabular*}{\textwidth}{@{\extracolsep{\fill}}llccccc@{}}
    \toprule
    \multicolumn{2}{@{}l}{\textbf{Travel-direction agreement (\%)}} & \multicolumn{5}{c@{}}{\textbf{Model}} \\
    \cmidrule(l){3-7}
    \textbf{Axis} & \textbf{Cue} & \textbf{Wan 2.2} & \textbf{SANA-V2} & \textbf{CogVideoX} & \textbf{SANA-WM} & \textbf{Lyra 2.0} \\
    \midrule
    Lateral & Camera sentence & 88 & 0 & 0 & 100 & 100 \\
     & Entropy asymmetry & 54 & 50 & 58 & 54 & 42 \\
     & Edge-density asymmetry & 62 & 50 & 58 & 54 & 50 \\
     & Near-depth asymmetry & 29 & 58 & 33 & 54 & 33 \\
     & Vanishing-point offset & 75 & 46 & 54 & 50 & 62 \\
     & DINOv3 saliency centre & 50 & 54 & 38 & 58 & 46 \\
     & Mirror reversal & 54 & 4 & 46 & 8 & 4 \\
     & Blur avoidance & 48 & 44 & 44 & 52 & -- \\
    \addlinespace[2pt]
    Vertical & Entropy, bottom vs.\ top & 20 & 27 & 46 & 0 & -- \\
     & Edge density, bottom vs.\ top & 60 & 41 & 46 & 0 & -- \\
     & Near depth, bottom vs.\ top & 20 & 27 & 46 & 0 & -- \\
    \addlinespace[2pt]
    Zoom & Open centre & 91 & 96 & 78 & 50 & 100 \\
    \addlinespace[2pt]
    Summary & Free lateral drift (px) & 2 & 30 & 13 & 1 & 0 \\
     & Free vertical drift (px) & 1 & 5 & 4 & 0 & 0 \\
     & Seed agreement (\%) & 73 & 90 & 75 & 94 & 99 \\
    \bottomrule
  \end{tabular*}
  \end{adjustbox}

  \vspace{6pt}

  \begin{adjustbox}{max width=\textwidth}
  \begin{tabular*}{\textwidth}{@{\extracolsep{\fill}}llccccc@{}}
    \toprule
    \multicolumn{2}{@{}l}{\textbf{Spearman $\rho$ for scene cues}} & \multicolumn{5}{c@{}}{\textbf{Model}} \\
    \cmidrule(l){3-7}
    \textbf{Axis} & \textbf{Cue} & \textbf{Wan 2.2} & \textbf{SANA-V2} & \textbf{CogVideoX} & \textbf{SANA-WM} & \textbf{Lyra 2.0} \\
    \midrule
     & Entropy asymmetry & $-$0.14 & $-$0.17 & 0.15 & 0.02 & $-$0.08 \\
     & Edge-density asymmetry & $-$0.03 & $-$0.17 & 0.09 & $-$0.07 & 0.07 \\
     & Near-depth asymmetry & $-$0.22 & 0.05 & $-$0.48 & $-$0.13 & $-$0.27 \\
     & Vanishing-point offset & 0.37 & 0.25 & 0.22 & 0.35 & 0.24 \\
     & DINOv3 saliency centre & $-$0.25 & $-$0.19 & $-$0.18 & $-$0.06 & $-$0.16 \\
    \addlinespace[2pt]
    Vertical & Entropy, bottom vs.\ top & $-$0.90 & $-$0.36 & $-$0.30 & -- & -- \\
     & Edge density, bottom vs.\ top & $-$0.50 & $-$0.24 & $-$0.12 & -- & -- \\
     & Near depth, bottom vs.\ top & $-$0.90 & $-$0.32 & $-$0.64 & -- & -- \\
    \addlinespace[2pt]
    Zoom & Open centre & $-$0.17 & 0.21 & 0.50 & -- & 0.50 \\
    \bottomrule
  \end{tabular*}
  \end{adjustbox}
\end{table*}

\subsection{Camera Motion Ratios and Direction Error}
\label{app:travel_metric}

\begin{figure}[!htbp]
  \centering
  \includegraphics[width=\linewidth,height=0.28\textheight,keepaspectratio]{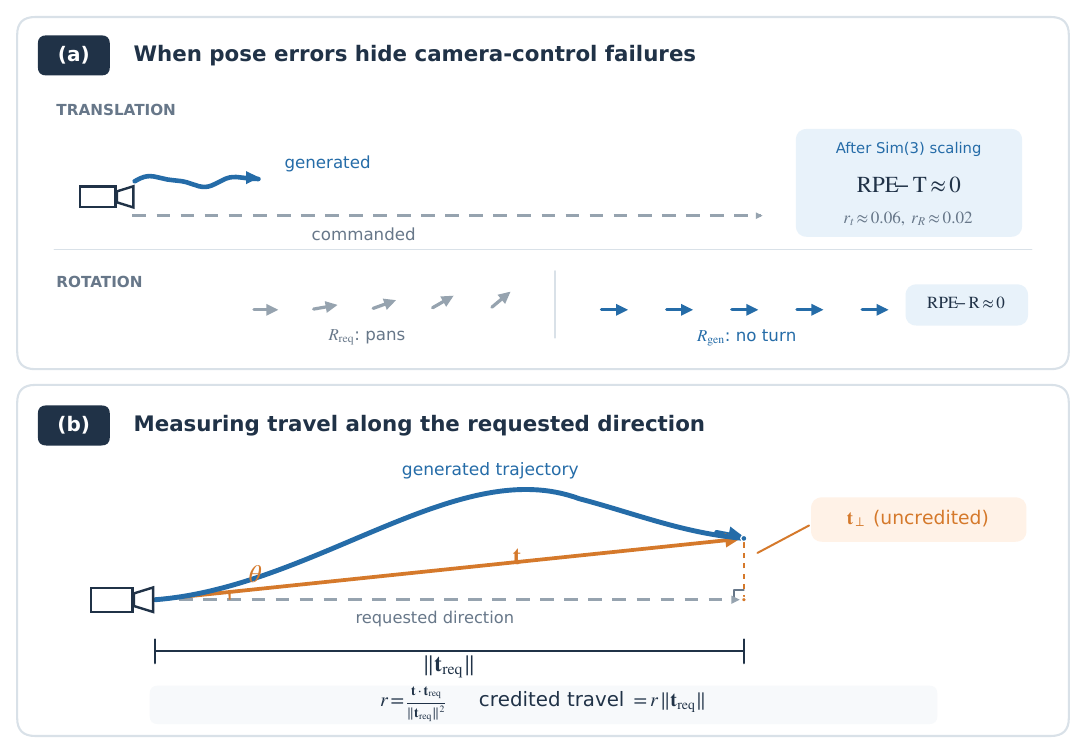}
  \caption{\textbf{Why net camera travel must be measured.}
  (a) Sim(3) fitting can rescale a nearly stationary trajectory
  and reduce translation error, while small per-step rotations
  can make a stationary view appear favorable under RPE-R.
  (b) Our unaligned metric-scale evaluation projects net camera
  displacement onto the requested direction. Off-axis travel
  does not contribute to the travel ratio and is measured
  separately through direction error $\theta$.}
  \label{fig:metric_diagram}
\end{figure}

\noindent \textbf{Motivation and metric scale.}
Relative pose error describes agreement between local camera
movements, but does not directly express how far the generated
camera travels toward the requested viewpoint. Sim(3)-aligned
RPE can further hide undershoot by rescaling the generated path,
as shown in Figure~\ref{fig:metric_diagram}. Our tabulated RPE
already uses unaligned trajectories. We additionally report
net displacement to quantify the final camera position.

Let $\mathbf{t}$ be the generated first-to-last camera
translation in the initial camera frame and
$\mathbf{t}_{\mathrm{req}}$ the corresponding requested
translation. The signed travel ratio and direction error are
\begin{equation}
\label{eq:supp_travel_ratio}
\begin{aligned}
r &= \frac{\mathbf{t}^{\mathsf T}\mathbf{t}_{\mathrm{req}}}
         {\lVert\mathbf{t}_{\mathrm{req}}\rVert_2^2},\\
\theta &= \cos^{-1}\!\left(
\frac{\mathbf{t}^{\mathsf T}\mathbf{t}_{\mathrm{req}}}
{\lVert\mathbf{t}\rVert_2\,
 \lVert\mathbf{t}_{\mathrm{req}}\rVert_2}\right).
\end{aligned}
\end{equation}
The ratio $r=1$ represents the requested displacement along
the target direction, $r=0$ no net travel along that axis,
$r<0$ travel in the opposite direction, and $r>1$ overshoot.
The projection does not credit perpendicular displacement.
For pure rotations, we compute the analogous signed ratio
from the first-to-last rotation vectors and use their axes
for the direction error. Translation and rotation components
are evaluated separately for mixed camera motions. The angular error is mathematically defined only when both motion vectors are nonzero. Near-zero travel is separately captured by the ratio metric.

\noindent \textbf{Aggregation and endpoint behavior.}
Table~\ref{tab:camprobe} reports mean angular direction error and median
$|r-1|$ over clips for travel and rotation separately.
We also examine the share of clips with $\theta<45^\circ$
and with ratios in $[0.75,1.25]$. Taking the absolute error
before aggregation prevents over- and undershoot from
canceling. The ratio measures first-to-last displacement,
not the length of the generated path. A curved trajectory
ending at the requested viewpoint can have the correct
endpoint ratio, while differences along the path remain
visible in unaligned RPE.

\begin{figure}[!htbp]
  \centering
  \includegraphics[width=\linewidth,height=0.31\textheight,keepaspectratio]{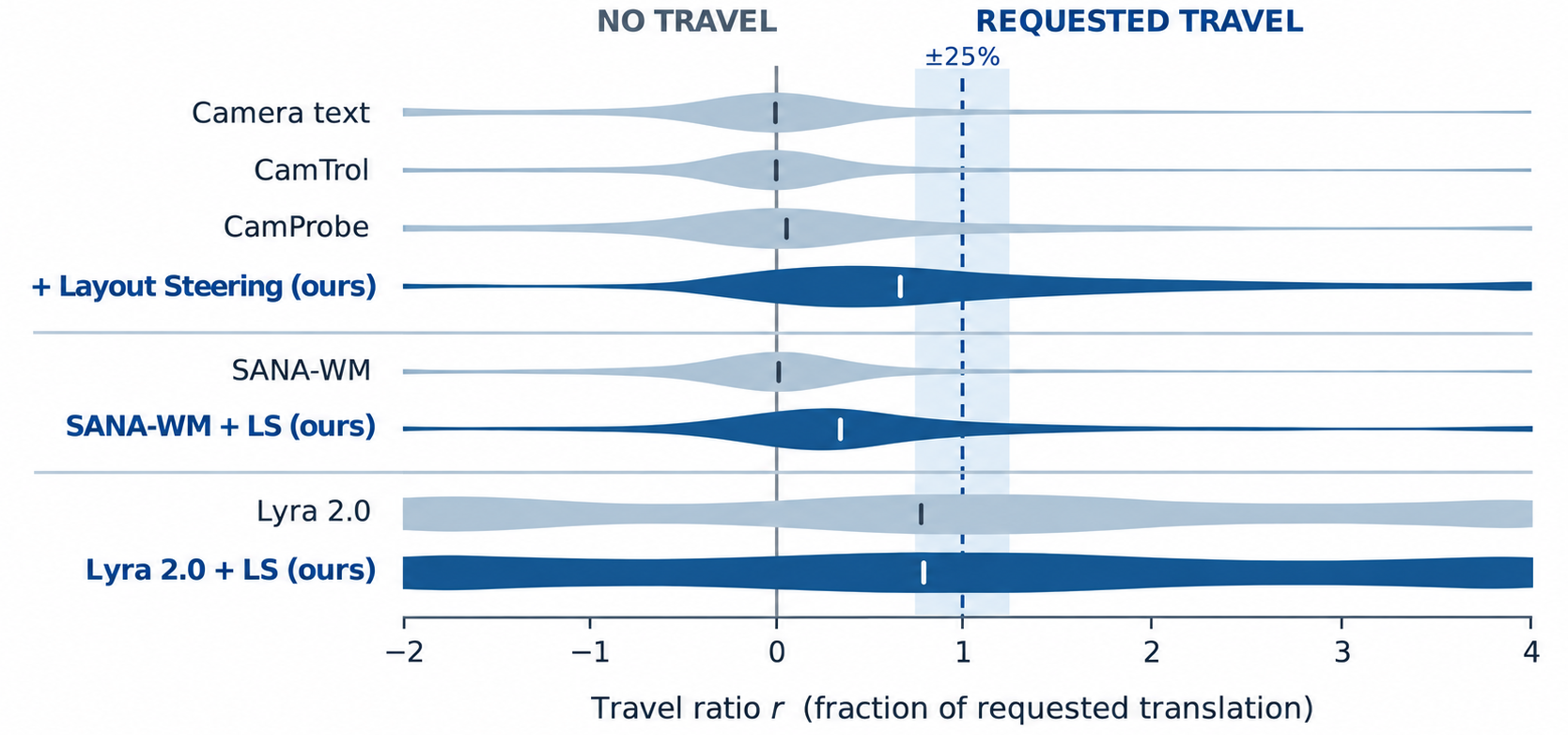}
  \caption{\textbf{Distribution of signed travel ratios.}
  Translational motions in Table~\ref{tab:camprobe} of the main paper are
  shown with median markers. The dashed line indicates
  requested travel ($r=1$), and the shaded band marks
  $r\in[0.75,1.25]$. The displayed range is clipped to
  $[-2,4]$. Wide and asymmetric distributions explain
  why median travel error and the fraction within
  $\pm25\%$ of the requested distance provide complementary
  summaries. Rotation ratios are evaluated separately.}
  \label{fig:r_dist}
\end{figure}

Figure~\ref{fig:r_dist} illustrates the variation in
signed travel ratios across the translational requests.
In particular, a moderate median absolute error does not
imply that most clips travel within $\pm25\%$ of the
requested displacement. Lyra~2.0 exhibits a broad
range that includes reversed travel and large overshoot.

\noindent \textbf{Independent image-space check.}
For selected examples, we compare the accumulated median
optical flow in each generated clip against its rendered
guide. On the larger 1.6\,m teaser trajectories, flow-based
and pose-based estimates agree within approximately 15
percentage points per displayed example. The shorter VBench
movements are closer to the pose-estimation noise floor.
Consequently, Table~\ref{tab:camprobe} uses jointly reconstructed camera
poses rather than optical flow. The optical-flow study of
spontaneous camera travel below is a separate within-model
analysis.

\section{Camera-Bias Probe}
\label{app:bias}

\noindent \textbf{Starting frames.}
The camera-bias probe uses 24 synthetic FLUX.3 Klein images
covering interiors, streets, architecture, and landscapes.
Each frame is generated from a distinct caption at
$1344\times768$ resolution with eight sampling steps and
one seed per caption. The same frames are used to initialize
all evaluated models. This avoids paired reference camera
trajectories, although it does not establish that the
starting images were absent from model training data.

\noindent \textbf{Models and generation settings.}
Wan~2.2 5B generates 81 frames at $1280\times720$ and 16\,fps,
with 50 denoising steps and guidance scale 3.5.
SANA-Video~2.0 5B generates 121 frames at $1280\times736$
and 24\,fps, with 50 steps, guidance scale 8, flow shift 12,
and motion score 20. CogVideoX~1.5 5B generates 81 frames
at $1360\times768$ and 16\,fps, with 50 steps and dynamic
guidance 6. All clips are approximately five seconds long.
Starting images are resized to each model's native resolution.
SANA-WM and Lyra~2.0 receive the requested camera trajectory
through their native conditioning interfaces.

\noindent \textbf{Camera requests and perturbations.}
We evaluate six camera conditions comprising no instruction,
a stationary camera, truck left, truck right, forward travel,
and rising travel. Text-conditioned requests also ask the
scene to remain static. Repeated conditions use five seeds,
while other conditions use one seed, with the available
numbers of clips reported in Table~\ref{tab:bias_full}.
We horizontally flip the starting images to test whether
the direction of travel reverses with scene composition.
We additionally blur the left or right 40\% of an image
using a 20-pixel Gaussian radius and a 64-pixel feathered
boundary to weaken local visual cues.

\noindent \textbf{Optical-flow travel.}
We resize generated clips to a width of 512 pixels and
accumulate frame-to-frame median horizontal and vertical
Farneb\"ack optical flow and mean flow divergence.
Under the camera-travel sign convention, positive lateral
travel denotes leftward camera movement and negative travel
denotes rightward movement. Positive vertical travel denotes
upward motion, while positive flow divergence denotes
forward travel. A requested direction is followed when
the measured sign agrees with the request. Stationary
lateral travel is defined as less than 10 pixels in magnitude.
These image-space measurements are separate from the metric
travel ratio in Section~\ref{sec:motion_metrics}.

\noindent \textbf{Scene-composition cues.}
We extract local Shannon entropy from $32\times32$ intensity
tiles, Canny edge density, relative inverse depth from Depth
Anything V2, DINOv3 ViT-B/16 patch saliency, and a
vanishing-point estimate obtained from smoothed depth.
Each cue defines a left-right asymmetry $(R-L)/(R+L)$ or
a horizontal position relative to image centre. Vertical
cues use corresponding top-bottom asymmetries. The
open-centre cue compares inverse depth in the centre and
periphery. We report the fraction of cases whose spontaneous
travel direction agrees with a cue, together with Spearman
correlation between cue strength and the observed travel.

\noindent \textbf{Probe findings.}
Table~\ref{tab:bias_full} reports travel under the six camera
requests, and Table~\ref{tab:bias_predictors} examines its
relationship with image composition. SANA-Video~2.0 exhibits
substantial spontaneous lateral drift, while Wan~2.2 responds
more reliably to opposing left and right requests. CogVideoX's
unprompted direction is more dependent on starting-frame
content. The trajectory-conditioned world models follow
requested directions more consistently, but their travel
distances still vary across scenes and seeds, as quantified
in Table~\ref{tab:bias} of the main paper.

\section{Compute and Runtime}
\label{app:overhead}

Layout Steering adds three operations to the frozen video prior.
A frozen image inpainting model extends the input image, depth-based
rendering and VAE encoding produce the guide, and a low-frequency
correction is applied during the first $K$ denoising steps.
Table~\ref{tab:overhead} reports measurements on one H100 for
an 81-frame Wan~2.2 5B clip at $1280\times704$ with 50 steps.
This configuration is distinct from the 49-frame CamProbe
benchmark. The steering correction itself adds less than
one second to 84 seconds of video sampling, while guide
preparation takes approximately 20 seconds.

\par\addvspace{6pt}
\noindent\begin{minipage}{\columnwidth}
  \makeatletter
  \def\@captype{table}
  \makeatother
  \centering
  \caption{\textbf{Per-window runtime on one H100.}
  Measurements use Wan~2.2 5B at $1280\times704$ with 81 frames
  and 50 denoising steps, not the 49-frame CamProbe benchmark.
  Canvas expansion with FLUX.1 Fill is not included in these measurements.}
  \label{tab:overhead}
  \footnotesize
  \setlength{\tabcolsep}{2pt}
  \renewcommand{\arraystretch}{1.12}
  \begin{tabular*}{\columnwidth}{@{\extracolsep{\fill}}>{\raggedright\arraybackslash}p{.52\columnwidth}>{\raggedright\arraybackslash}p{.23\columnwidth}r@{}}
    \toprule
    \textbf{Stage} & \textbf{Runs} & \textbf{Time} \\
    \midrule
    Video prior, one window & per window & 84\,s \\
    Steering hook ($K$ low-pass corrections) & per window & $<$1\,s \\
    Guide (depth, reprojection, VAE encoding) & per window & 20\,s \\
    \bottomrule
  \end{tabular*}
\end{minipage}\par\addvspace{6pt}

These measurements exclude the initial FLUX.1 Fill outpainting cost and therefore do not represent end-to-end runtime. A canvas can be reused
across multiple requested trajectories from the same starting
image. In a chained tour, the final frame of each video segment
is expanded again to construct a new guide for the next segment.

\onecolumn
\section{Additional Visual Results}
\subsection{Additional Qualitative Comparisons}
Figures~\ref{fig:bench_full} and~\ref{fig:bench_full_trained}
extend the single-window comparisons in Figure~\ref{fig:lay_results} of the
main paper across the same camera motion categories. The
examples are intended as visual complements to the
quantitative tables, not additional benchmark populations.
\noindent\begin{minipage}{\textwidth}
  \makeatletter
  \def\@captype{figure}
  \makeatother
  \centering
  \includegraphics[width=0.75\textwidth,height=0.39\textheight,keepaspectratio]{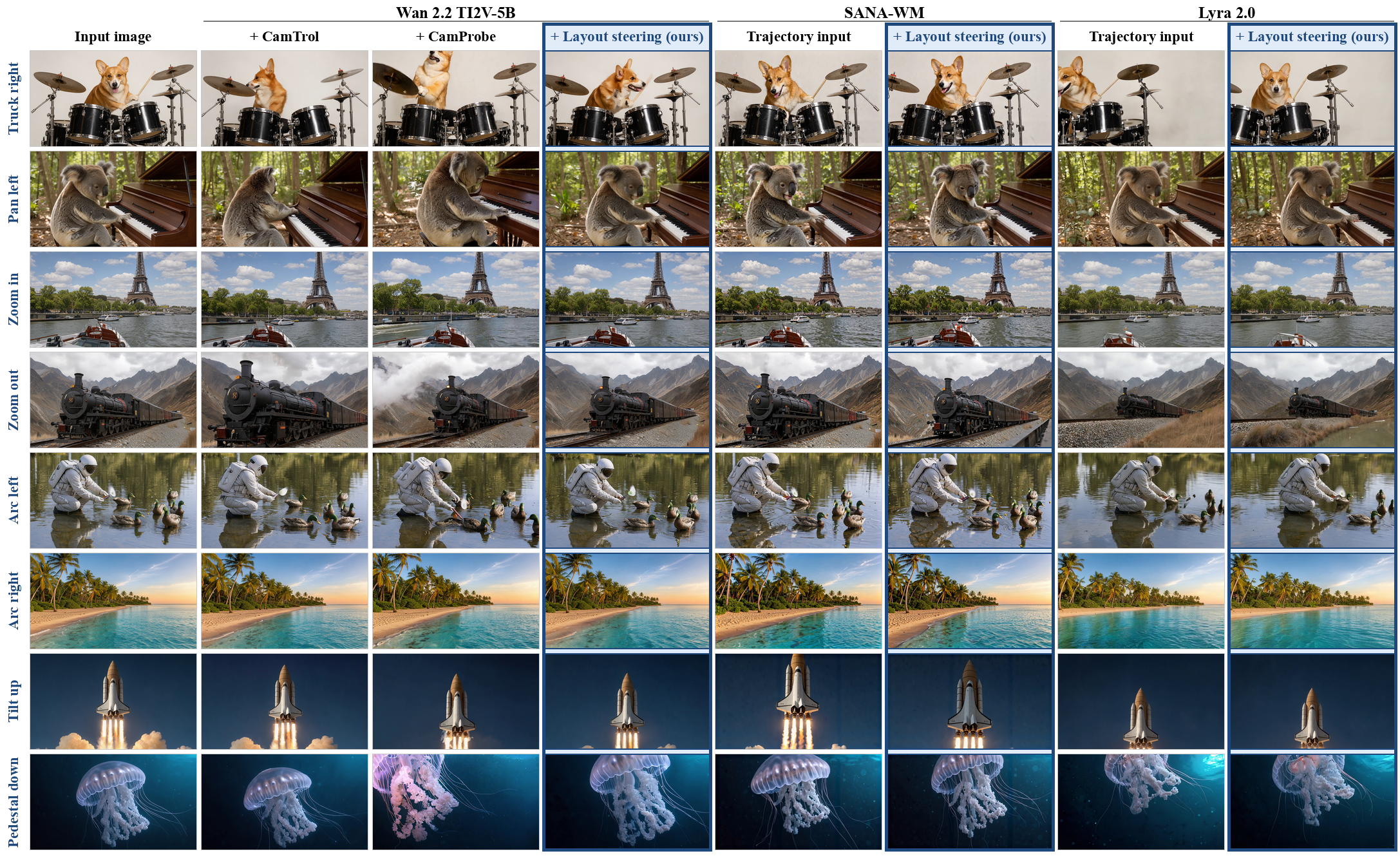}
  \caption{\textbf{Qualitative camera control across eight requested motions.}
  Additional CamProbe-protocol examples complement Figure~\ref{fig:lay_results} of
  the main paper. CamTrol can transfer guide-rendering artifacts
  when restarting sampling. Layout Steering corrects coarse
  spatial structure while the video prior synthesizes fine detail.}
  \label{fig:bench_full}
\par\vspace{8pt}
  \centering
  \includegraphics[width=0.75\textwidth,height=0.39\textheight,keepaspectratio]{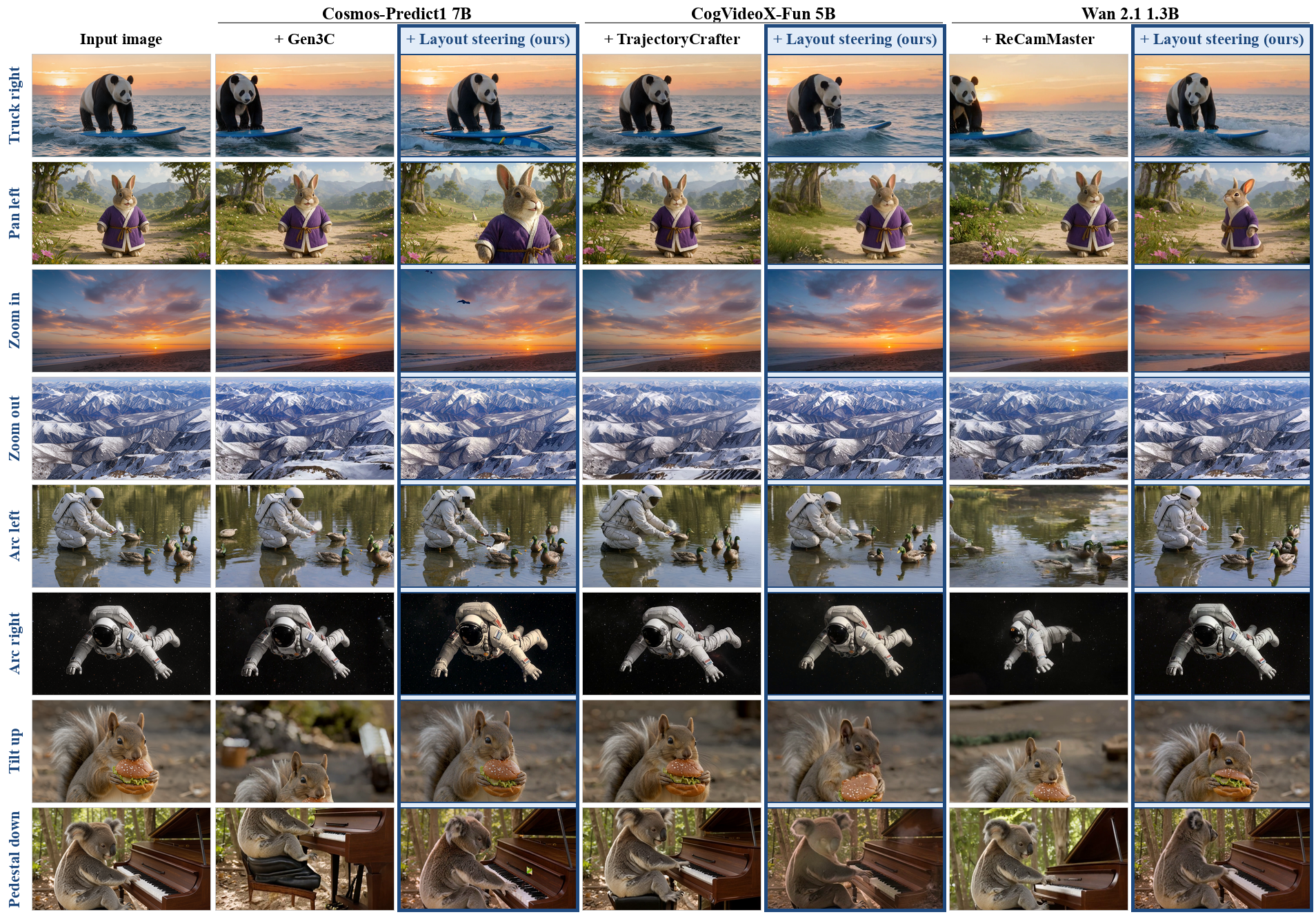}
  \caption{\textbf{Comparison with trained camera-control methods.}
  Additional results for the learned camera-control baselines
  and Layout Steering in Table~\ref{tab:camprobe} of the main paper. Each group
  uses the same starting image and requested camera motion.}
  \label{fig:bench_full_trained}
\end{minipage}\par

\subsection{Camera-Bias Probe Starting Images}
\noindent\begin{minipage}{\textwidth}
  \makeatletter
  \def\@captype{figure}
  \makeatother
  \centering
  \includegraphics[width=\textwidth,height=0.86\textheight,keepaspectratio]{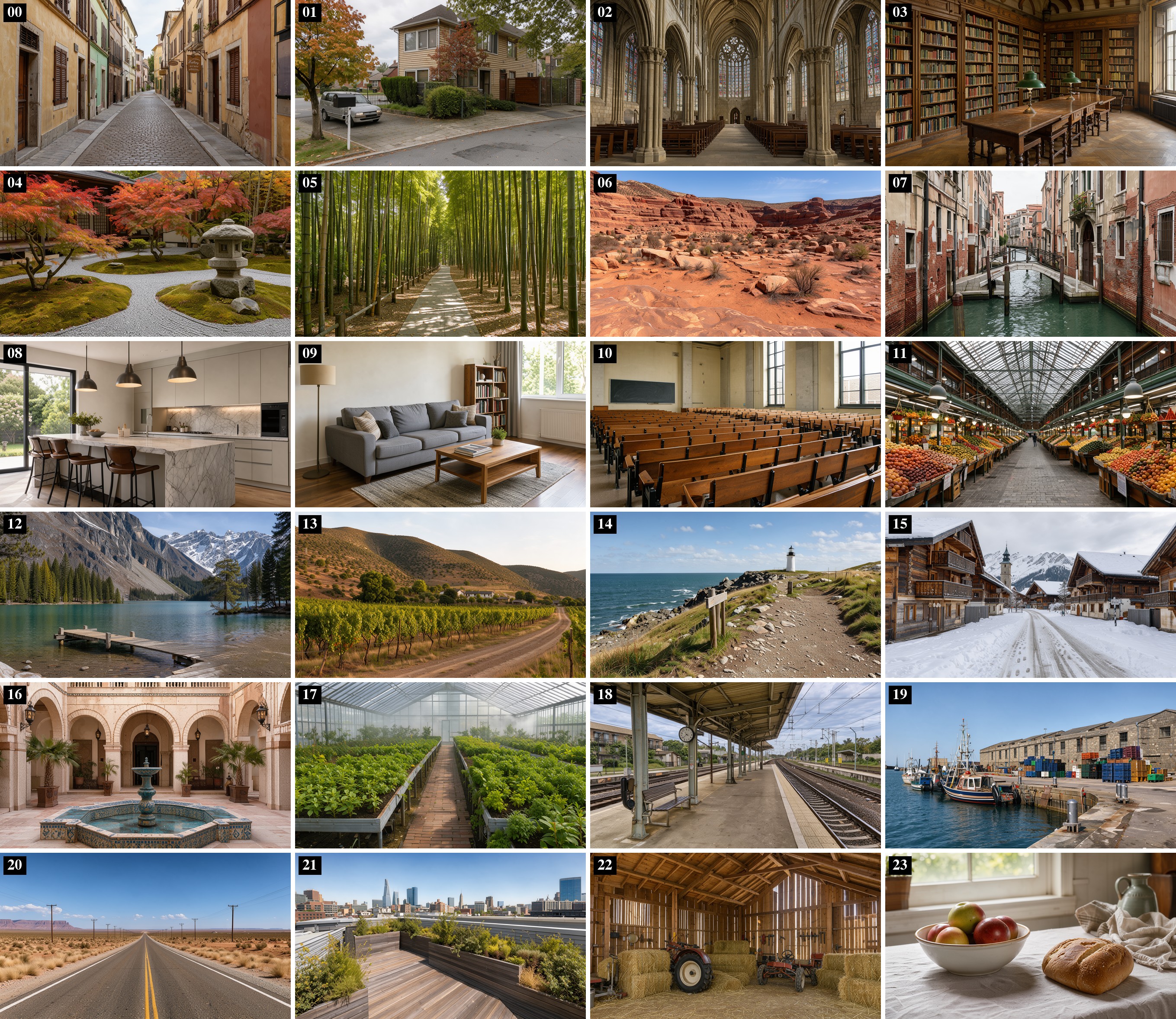}
  \caption{\textbf{Starting images for the camera-bias probe.}
  The 24 FLUX.3 Klein images are generated from distinct captions
  at $1344\times768$ resolution, ordered by index in row-major
  order. Each image is used without paired ground-truth camera
  trajectories for the experiments in Table~\ref{tab:bias} of the main paper.}
  \label{fig:frames}
\end{minipage}\par

\end{document}